%% file: main.tex
\PassOptionsToPackage{hyphens}{url}
\documentclass[10pt,twocolumn,letterpaper]{article}

\usepackage{times}
\usepackage{tabularx}
\usepackage{multicol}
\usepackage[bookmarks=true]{hyperref}
\usepackage{wrapfig}
\usepackage{diagbox}
\usepackage{times}
\usepackage{epsfig}
\usepackage{graphicx}
\usepackage{amsmath}
\usepackage{amssymb}
\usepackage{multicol}
\usepackage{wrapfig}
\usepackage{lscape}
\usepackage{caption}
\usepackage{booktabs}
\usepackage{tabularx}
\usepackage{array}
\usepackage{multirow}
\usepackage{graphicx}
\usepackage{subcaption}
\usepackage{setspace}
\usepackage[symbol*]{footmisc}
\usepackage{algorithm}
\usepackage{algpseudocode}
\usepackage{amsmath}
\usepackage{graphicx}
\usepackage{subcaption}  
\usepackage[margin=1in]{geometry}
\usepackage[accsupp]{axessibility} 
\setfnsymbol{wiley}
\usepackage{xspace}

\usepackage[colorinlistoftodos]{todonotes}
\usepackage{cvpr}              
\usepackage{xcolor}
\input{preamble}

\definecolor{cvprblue}{rgb}{0.21,0.49,0.74}

\def\paperID{18368} 
\def\confName{CVPR}
\def\confYear{2026}

\title{Expanding Spatial and Temporal Context for Robotic Imitation Learning With Scene Graphs}

\author{
  Jianing Qian$^{1}$,
  Qinhe Peng$^{1}$\thanks{Equal contribution.} ,
  Emmanuel Panov$^{2}$\footnotemark[1] ,
  Leonor Fermoselle$^{2}$,\\
  Dinesh Jayaraman$^{1}$\thanks{Equal advising.}    ,
  Bernadette Bucher$^{3}$\footnotemark[2]    ,
  Tarik Kelestemur$^{2}$\footnotemark[2] \\
  $^{1}$University of Pennsylvania,
  $^{2}$RAI Institute,
  $^{3}$University of Michigan\\
  \texttt{\{jianingq,pengqh20,dineshj\}@seas.upenn.edu},\\
  \texttt{\{epanov,lfermoselle,tkelestemur\}@theaiinstitute.com},\\
  \texttt{bucherb@umich.edu}
}

\begin{document}
\maketitle
\input{sec/0_abstract}    
\input{sec/1_intro}
\input{sec/2_related}
\input{sec/3_method}

\input{sec/4_experiments}
\newpage
\section*{Acknowledgments}
This work was supported in part by the National Science Foundation (NSF) under CAREER Award \#2239301 and SLES Award \#2331783, the Office of Naval Research (ONR) under Grant N00014-22-1-2677, and DARPA under Grant HR00112490421 (TIAMAT).

{
    \small
    \bibliographystyle{ieeenat_fullname}
    \bibliography{main}
}
\input{sec/5_suppl}
\end{document}

%% file: preamble.tex
\usepackage{caption}

%% file: sec/0_abstract.tex
\begin{abstract}
Imitation learning enables robots to learn how to execute tasks via observation. However, real-world environments like homes and offices are often severely partially observed due to their large spatial scales. In addition, many tasks involve executing a series of subtasks requiring autonomous robots to reason over extended time horizons. 
To address these challenges, we propose using scene graphs as an explicit and structured memory mechanism in imitation learning. By maintaining a dynamic scene graph that captures object-centric relationships and their evolution over time, our method allows the agent to retain relevant historical context during task execution to efficiently reason over incrementally accrued scene information. Our experiments on simulated mobile manipulation and real-world tabletop manipulation demonstrate that our approach substantially improves policy performance, particularly in settings that demand long-term reasoning and robust generalization under partial observability. Code and videos:https://pengqinhe.github.io/Scene-Graph-CVPR2026-website/.
\end{abstract}

%% file: sec/1_intro.tex
\section{Introduction}
\label{sec:intro}
Imitation-learned policies have shown great promise for executing complex physical robotic skills, but largely over small spatial scales and short time horizons~\cite{zhang2018deep,dasari2020transformers,florence2022implicit,zhang2021performer}. At larger spatial and temporal contexts, their performance frequently deteriorates~\cite{jang2022bcz,mandlekar2021what}. A key source of difficulty in such tasks is partial observability. Consider, for example, a mobile manipulation task where it needs to move a chair to a different room. If a robot only has access to onboard sensor information during task execution, it is commonly not possible to fully observe a scene while executing actions. Nevertheless, performing such tasks well often requires the robot to reason over information that is not currently in its field of view.

This motivates maintaining some form of memory for robotic agents operating in partially observed settings.  What might such a memory look like? Modern computer vision approaches permit building and maintaining comprehensive metric maps aggregating information from all previous views of the scene, but such an overcomplete memory creates new issues for imitation policy learning: how should such large maps be processed in a neural policy, and how can we train such a policy sample-efficiently? 

We propose to construct and maintain a novel scene memory representation for long spatial and temporal context imitation learning. When a policy is conditioned on this representation, it can make decisions based on all task-relevant information in the scene, even if it was observed in the distant past.
We draw inspiration from the ample evidence to suggest that humans instead maintain abstract task-relevant \textit{semantic} memories when accomplishing long-horizon tasks~\cite{tulving1972episodic,kumar2021semantic,he2024human,doerig2024visualrepresentationshumanbrain,khan2025survey,plos2024decoding}.  
Mirroring this, we use scene graphs to store scene information compactly for large environments in a manner that can be easily updated with dynamic environment changes. Scene graphs sparsely represent critical scene information over large spaces, even up to the scale of multi-floor buildings.  Updating a scene graph from movement in the environment is straightforward since scene information is clearly discretized and organized in a graph structure. Furthermore, designing the scene graph to only track task-relevant information can enable efficient learning and execution of targeted actions even in large, cluttered scenes. 
However, no prior work directly leverages a scene graph as a model input in imitation learning. We fill this gap: we propose an approach to construct and dynamically update task-relevant scene graphs to provide the necessary spatial and temporal context for training and executing imitation learning neural policies in partially observed settings.


%% file: sec/2_related.tex
\section{Related Work}
\label{sec:related_work}
We begin by summarizing prior efforts on generating scene graphs using vision foundation models. We then review approaches that incorporate scene graphs into robot policy learning. Finally, we discuss imitation learning methods that introduce memory mechanisms to improve performance on long-horizon tasks.

\noindent\textbf{Semantic Scene Graph with Vision Foundation Models.}
Scene graphs~\cite{Xu2017SceneGG, Li2017SceneGG, Yang2018GraphRF, Lin2020GPSNetGP} have grown popular as a structured and interpretable image representation. In a scene graph, salient objects are represented as graph nodes, and edges represent their spatial relationships. The concept of the scene graph can naturally be extended from 2D to 3D, where input to the scene graph generation pipelines is no longer restricted to image observations, but also 3D maps of an environment. Previous works~\cite{Gay2018VisualGF,Armeni20193DSG,Zhang2021Knowledgeinspired3S,Wu2021SceneGraphFusionI3} introduce methods to build 3D semantic scene graphs, either from videos, 3D mesh models, or 3D point clouds of the visual environment.
Given the recent advances of vision-language foundation models~\cite{SAM,dino,Oquab2023DINOv2LR,Liu2023GroundingDM,Zhang2023RecognizeAA,Radford2021LearningTV}, many works propose to leverage their generalization ability in new environments and ground the output 2D representations from those models in 3D point clouds or maps.~\citet{Chang2023ContextAwareEG,Jatavallabhula2023ConceptFusionOM,Gu2023ConceptGraphsO3,chang2025ashitaautomaticscenegroundedhierarchical} propose frameworks to generate semantic 3D scene representations by prompting VLMs.

Most related to our approach is CLIO~\cite{Maggio2024ClioRT}, which builds on a hierarchical 3D scene graph~\cite{Hughes2023FoundationsOS} to provide a framework to infer the appropriate granularity and sub-graph given user-input task instructions. Unlike CLIO, which only demonstrates simple object picking tasks, we focus on how to enable long-horizon task execution with a 3D semantic scene graph. 
Many other prior works also utilize scene graphs or object-centric representations to improve VLA models~\cite{Bendikas2025FocusingOW}, guide long-horizon task planning~\cite{Wang2025ODYSSEYOQ,Ni2023GRIDSI,Yan2024DynamicO3,Yang2025InterleavedLA,ashton2025helios}, and actively explore the environment or navigate~\cite{Jiang2024RoboEXPAS,Devarakonda2024OrionNavOP,Yokoyama2023VLFMVF,Ravichandran2021HierarchicalRA,raychaudhuri2024zero}. Unlike us, none of them show how modern neural imitation learning architectures could be modified to directly ingest dynamically maintained scene graphs as inputs. Similar to us, ~\cite{Sieb2019GraphStructuredVI} also constructs scene graphs to aid visual imitation learning. However, they use scene graphs to map human demonstrations to robot ones,  while we directly encode and input the scene graph to the downstream diffusion policies.

\noindent\textbf{Imitation Learning with Memory.}
Imitation learning methods such as Diffusion Policy (DP)~\cite{chi2023diffusion}, Action Chunking Transformers (ACT)~\cite{zhao2023learning}, and several variants of these methods~\cite{zhu2023learning, ze20243d, goyal2024rvt} over the years have shown that we can learn dexterous and multi-modal tasks with a few hundred demonstrations. Later, by scaling the data and the network capacity~\cite{saxena2025matters, intelligence2025pi_, yan2025maniflow}, we see that we can learn multi-task policies that are conditioned on the language or image goals. However, these models are still not able to handle long-horizon and large spatial information, as the majority of these works address tabletop manipulation, where partial observability is not a major issue. These challenges are particularly important for mobile manipulation, where the robot does not have an immediate view of its full environment. To this end,~\cite{torne2025learning} introduced a variant of DP that targeted long-context visual imitation learning. They introduce an auxiliary loss to the network to avoid learning spurious correlations in the long-context data. On the other hand, \cite{liu2024enabling} learns to incorporate state information by incorporating ControlNet. The efficient scene memory representation in our approach encodes both large-scale spatial and temporal context as input representations to DP. Critically, our memory representation is explicitly constructed, not learning over spatial information and time.


%% file: sec/3_method.tex
\section{Method}
\label{method}

\begin{figure*}[t]
    \centering
    \includegraphics[width=0.7\linewidth]{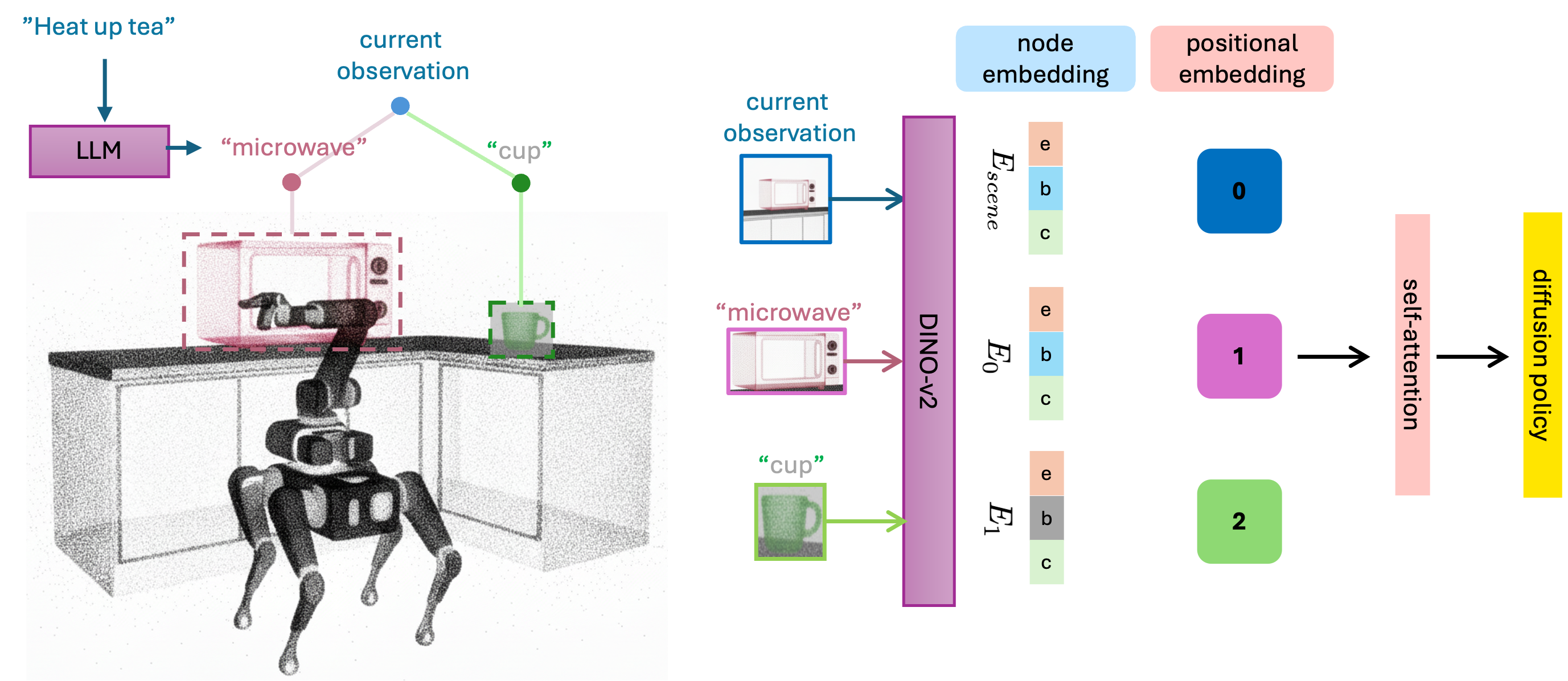}
    \caption{We maintain a scene graph where each task-relevant object is represented as a node 
\( n_i = [e_i, b_i, c_i] \), consisting of a learned visual embedding \(e_i\), its 2D bounding box \(b_i\) when visible, 
and its 3D centroid \(c_i\) in a world-aligned frame. Using a natural-language task description, large language models 
and vision foundation models identify and track relevant objects, allowing the graph to be incrementally updated as the robot interacts with the environment. 
This provides a compact, memory-efficient representation suitable for downstream imitation-learning policies. 
In the figure, the microwave is currently in view, while the cup is out of view (shown as a grayed-out area). 
We update \(b_i\) only when the object is visible.}
    \label{fig:concept_figure}
    \vspace{-1.5em}
\end{figure*}
We study the problem of imitation learning in partially observed environments. Our training dataset consists of $M$ expert demonstrations, $D = \{\tau_i\}_{i=1}^M$, where each trajectory $\tau_i$ is a sequence of observation-action pairs $\tau_i = [(I_0,a_0),... (I_T,a_T)]$ where $I$ is an RGB-D image. A key challenge in our setting is that the observation at each timestep, $I_t$, does not fully capture the underlying state $s_t$ of the environment.

This partial observability is a common and challenging scenario in robot learning. For example, on a mobile manipulation platform, visual observations are typically gathered from body-mounted cameras with a limited field-of-view that see only a very small portion of the relevant environment, such as a home or an office building. Even in table-top manipulation, wrist-mounted cameras are popular as they abstract away many details of the robot morphology, enable fast data collection, and induce robustness to extraneous information. However, in both cases, the robot must contend with incomplete and noisy sensory observations, making robust decision-making difficult.

Consider a mobile robot performing indoor household tasks, such as tidying a room or heating a cup of tea. With onboard sensors and limited perceptual coverage, the robot must act based on partial observations, without direct access to the full environment state. These settings are naturally modeled as Partially Observable Markov Decision Processes (POMDPs), where the agent must infer the hidden state of the world from a history of past observations and actions.

In principle, optimal policies $\pi^*$ in these partially observed settings could require access to the entire observation-action history. However, learning such a full history-dependent policy is typically intractable, especially in long-horizon tasks where the observation history grows rapidly and extracting relevant information becomes increasingly difficult~\cite{jiang2019value,dong2022simple,lu2023bitbybit,sinha2023agentstate}.

\subsection{Task-Driven Scene Graph}

To capture historical information compactly, we propose a structured memory representation in the form of a dynamically updated task-relevant scene graph, designed to support effective robotic decision-making under partial observability.

For our purposes, a scene graph refers to a graph-based abstraction in which task-relevant objects are represented as a set of nodes $N=\{n_i\}=\{[e_i,b_i,c_i]\}$, each consisting of an embedding $e_i$, its 2D bounding box coordinates $b_i$ (if it is currently in view, empty otherwise), and its 3D centroid $c_i$ in a world coordinate frame (anchored through odometry in mobile manipulation, or just attached to the robot base in tabletop manipulation). Importantly, while we do not explicitly parameterize edges as separate learnable entities, relational structure is implicitly encoded through shared coordinate frames and joint processing of node attributes. In particular, object centroids $c_i$ define a common spatial reference frame, allowing pairwise relationships (e.g., relative position and distance) to be inferred directly from node features. 

To identify and track potentially task-relevant objects, we leverage large language models and vision foundation models based on a concise natural language description of the task, as adopted from~\cite{hodor}. As the task unfolds, we incrementally update and maintain the scene graph to reflect the current state of the environment in a task-aware and memory-efficient way suitable as the state representation input to a neural imitation policy.
We discuss these steps in detail in the following paragraphs.

\noindent\textbf{Task-Relevant Object Enumeration.}
In realistic indoor environments, exhaustively identifying and tracking all objects is computationally infeasible and unnecessary. To build a scene graph that is both compact and effective for policy learning, we focus on identifying only the task-relevant objects based on the provided task description.

Given a natural language task instruction, we use a large language model (e.g., GPT-5) to extract a list of noun entities (details in Sec. ~\ref{sec:supp-graph}) that are likely to be important for the task. Once this list of task-relevant entity words is generated, we look to localize and track all objects matching any item on that list, as we describe below.
\\
\noindent\textbf{Scene Graph Construction and Maintenance.}
For the first image frame $I_0$ in an episode, we use Grounding DINO~\cite{Liu2023GroundingDM} to detect and localize objects matching the task-relevant entity list, producing a set of segmentation masks $M_{t=0} = \{m_0^j\}$. These detections initialize the nodes in the scene graph.

\textbf{Node Representation.}
At any time step $t$, given segmentation masks $M_t = \{m_t^j\}$ for the current observation, we represent each object as a node by extracting object-centric features. For each mask $m_t^j$, we compute an appearance embedding $e_j$ by average-pooling DINO-v2~\cite{Oquab2023DINOv2LR} features within the masked region. We also extract a tight 2D bounding box $b_j$ and estimate a 3D centroid $c_j$ via depth backprojection using camera parameters and odometry. This yields a node representation $n_j = \{e_j, b_j, c_j\}$ capturing both semantic and spatial information.

\textbf{Temporal Maintenance.}
To maintain the scene graph over time, we combine tracking of existing nodes with detection of newly observed objects. For each node already present in the graph, we initialize a visual tracker~\cite{Cheng2022XMemLV}, enabling persistent identity across frames. When an object moves out of view, its 2D bounding box $b_i$ is set to zero while its 3D centroid $c_i$ is retained, assuming no movement unless re-observed. When the object reappears, both $b_i$ and $c_i$ are updated accordingly.

Since the environment is only partially observed, new task-relevant objects may appear over time. To account for this, we run object detection in parallel at each time step. Each newly detected candidate $n_j$ is compared against existing nodes $n_i$ by computing mask overlap. If no existing node exceeds a threshold $\tau_m$, the detection is initialized as a new node and added to the scene graph.

This design differs from a flat set of object tokens in two key ways. First, nodes maintain persistent identities across time and are incrementally updated, forming a temporally consistent structure. Second, spatial relationships between objects are preserved through their shared coordinate representation, enabling relational reasoning without requiring explicitly parameterized edges.

See Algorithm~\ref{alg:scene-graph} for a detailed pseudocode of the scene graph update procedure.

\subsection{Policy Learning}
We now encode the scene graph as input to a neural diffusion policy. 

\noindent\textbf{Scene Graph-Conditioned Transformer Policy.}
Given the set of task-relevant object nodes \(N = \{n_i\} = \{[e_i, b_i, c_i]\}\), we construct a structured representation of the current scene and use it to condition a transformer-based policy.

\textbf{Scene and Object Encoding.}
For each observation \(I\), we first extract a global scene representation \(e_{\text{scene}}\) using the CLS token of a DINO-v2 ViT encoder. We associate this with a scene-level node defined by a bounding box \(b_{\text{scene}} = [0., 1., 0., 1.]\) in normalized image coordinates and a 3D coordinate \(c_{\text{scene}} = [0., 0., 0.]\) representing the origin of the global frame.

For each object node \(n_i \in N\), we compute its visual embedding \(e_i\) as described earlier. The geometric attributes, including the 2D bounding box \(b_i\) and 3D centroid \(c_i\), are encoded via small MLPs. The final node embedding is given by
\[
E_i = [e_i, \mathrm{MLP}(b_i), \mathrm{MLP}(c_i)].
\]
Similarly, the scene-level embedding is
\[
E_{\text{scene}} = [e_{\text{scene}}, \mathrm{MLP}(b_{\text{scene}}), \mathrm{MLP}(c_{\text{scene}})].
\]

\textbf{Scene Graph Representation.}
The full scene graph representation is given by the set of node embeddings
\[
R_{sg} = [E_{\text{scene}}, E_1, \ldots, E_i, \ldots],
\]
where \(E_{\text{scene}}\) serves as a root node and \(E_i\) correspond to task-relevant objects. This representation preserves object identity and spatial structure, while enabling relational reasoning through joint processing of all nodes.

\textbf{Transformer Encoding.}
We input \(R_{sg}\) as a sequence of tokens to a transformer encoder. Positional embeddings are used to encode the structural roles of nodes (e.g., a distinct embedding for the scene-level root). The transformer processes all tokens jointly via self-attention, allowing interactions between objects and the global scene context. The resulting features are aggregated through an MLP head to produce a compact scene representation, which conditions the downstream policy.

In our current tasks, the scene graph typically forms a shallow hierarchical structure (i.e., a scene-level root and object nodes). Despite this simplicity, it captures the object-centric state and spatial relationships necessary for decision-making.

\noindent\textbf{Imitation Learning with Diffusion.}
We use the scene graph representation \(R_{sg}\) to condition a diffusion-based policy that generates short-horizon action sequences.

During training, expert actions are corrupted using a forward diffusion process, and a denoising model \(\epsilon_\theta(a_k, R_{sg}, k)\) is trained to predict the injected noise under the DDPM objective:
\begin{align*}
\mathcal{L}(\theta) = 
\mathbb{E}_{\mathbf{a}_0, \, k \sim \mathcal{U}(1,K), \, \boldsymbol{\epsilon} \sim \mathcal{N}(0, \mathbf{I})}
\left[
\left\|
\boldsymbol{\epsilon} - 
\epsilon_{\theta}(a_k, R_{sg}, k)
\right\|^{2}
\right].
\end{align*}

At test time, the policy initializes from Gaussian noise and iteratively applies reverse diffusion (using a DDIM sampler) to produce a coherent action sequence conditioned on the current scene graph.

\begin{algorithm}[t]
\caption{Scene Graph Maintenance Pseudocode}
\label{alg:scene-graph}
\begin{algorithmic}[1]
\Require RGB image $I_{\text{rgb}}$, depth image $I_{d}$, camera intrinsics $E_i$, camera extrinsics $E_e$, task entities $\mathcal{N}_{\text{task}}$, overlap threshold $\tau_m$
\Ensure Scene graph $\mathcal{V}$
\State \textbf{Function} GenerateSceneGraph($I_{\text{rgb}}, I_d, E_i, E_e, \mathcal{N}_{\text{task}}, \tau_m$)
    // update scene graph at every timestep t
    \For{$v_i \in \mathcal{V}$}
        \If{\Call{IsVisible}{$v_i, I_{\text{rgb}}$}}
            \State $(e_i,b_i, c_i) \gets$ \Call{UpdateFromTracker}{$v_i$}
        \Else
            \State $b_i=(0,0,0,0)$, keep $c_i$ unchanged
        \EndIf
    \EndFor
    \State $M \gets$ \Call{DetectObjects}{$I_{\text{rgb}}, \mathcal{N}_{\text{task}}$}
    \For{$m^j \in M$}
        \State $v_i \gets$ \Call{OverlapWithExisting}{$m_j, \mathcal{V}, \tau_m$}
        \If{\textbf{not} $v_i$}
            \State $b_j \gets$ \Call{GetBBox}{$m^j$}
            \State $e_j \gets$ \Call{ExtractDINOFeatures}{$I_{\text{rgb}}, b_j$}
            \State $c_j \gets$ \Call{Compute3DCentroid}{$I_d, b_j,E_i, E_e$}
            \State $\mathcal{V} \gets \mathcal{V} \cup \{(e_j, b_j, c_j)\}$
        \EndIf    
    \EndFor
    \State \Return $\mathcal{V}$
\end{algorithmic}
\end{algorithm}

%% file: sec/4_experiments.tex
\section{Experiments}
\label{experiments}
In this section, we evaluate our method for learning long-horizon imitation policies under partial observability. We address three key questions: (1) Does our method outperform prior imitation learning baselines for long-horizon tasks? (2) How well does it generalize across diverse tabletop and mobile manipulation tasks? (3) How important are the individual components of our approach?

To answer these, we conduct experiments in both simulation and real-world settings and comprehensively evaluate performance.

\subsection{Simulated Mobile Manipulation}
\begin{figure}[t]
    \centering
    \begin{subfigure}{0.8\linewidth}
        \centering
        \includegraphics[width=\linewidth]{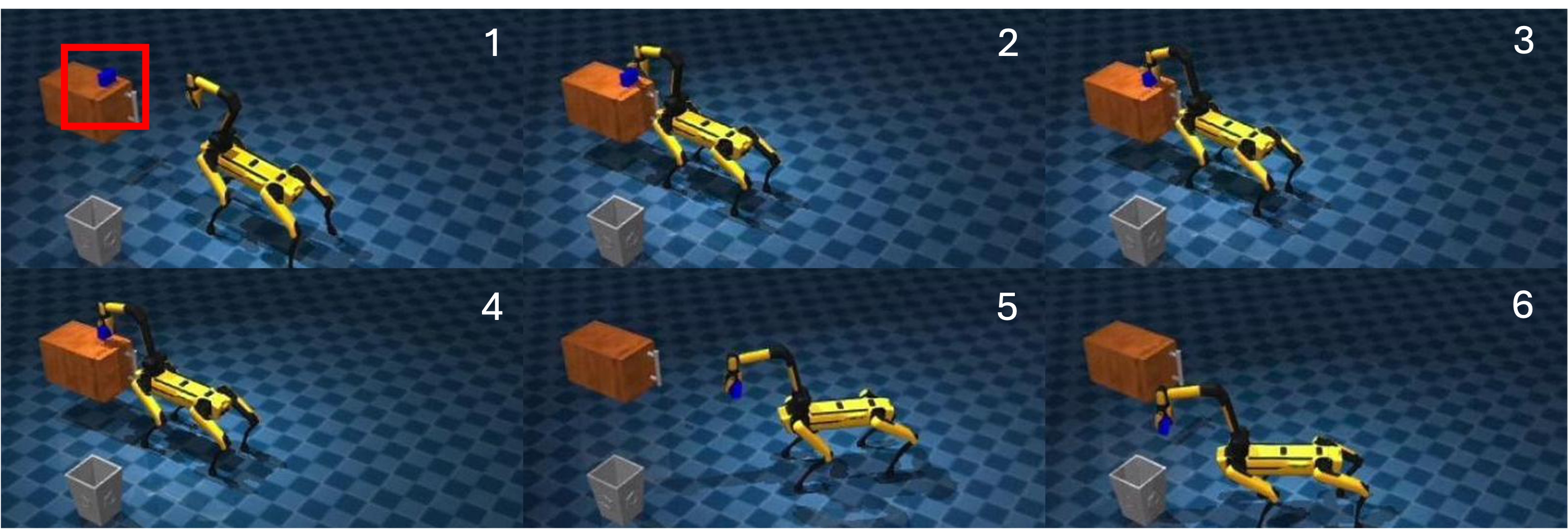}
        \caption{Task 1: Throw Trash}
        \label{fig:image_a}
    \end{subfigure}
    \vspace{0.5em} 

    \begin{subfigure}{0.8\linewidth}
        \centering
        \includegraphics[width=\linewidth]{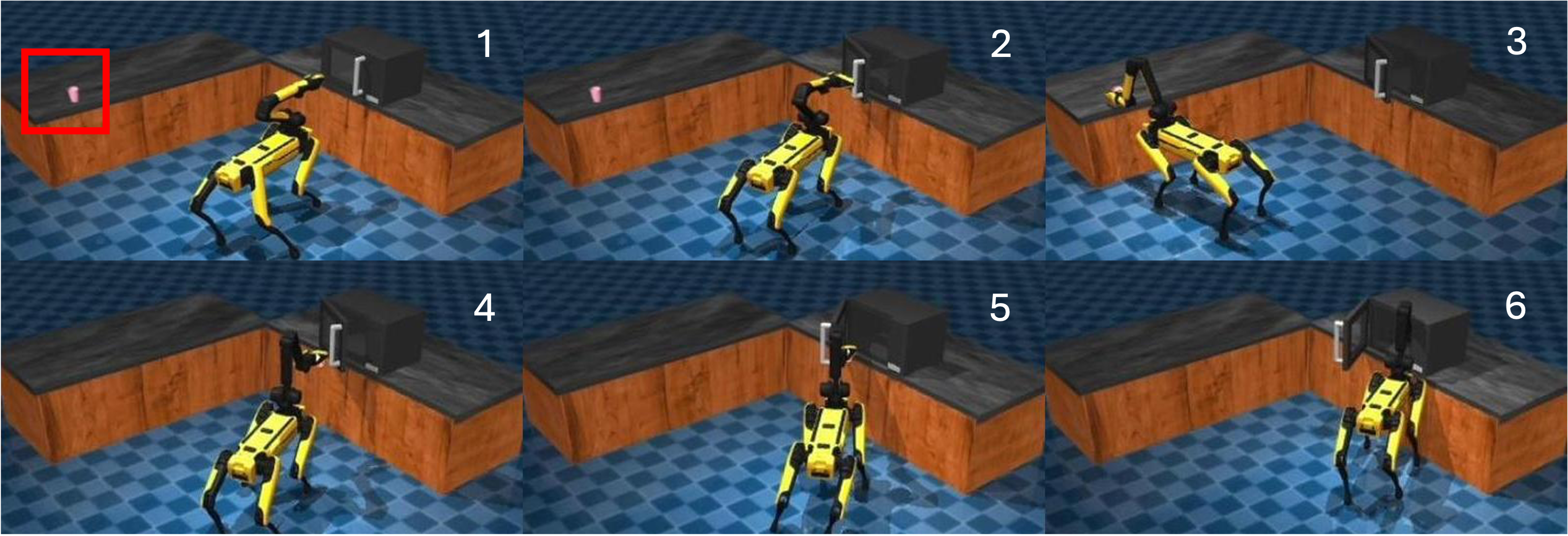}
        \caption{Task 2: Heat-up Tea}
        \label{fig:image_b}
    \end{subfigure}
    \vspace{0.5em}

    \begin{subfigure}{0.8\linewidth}
        \centering
        \includegraphics[width=\linewidth]{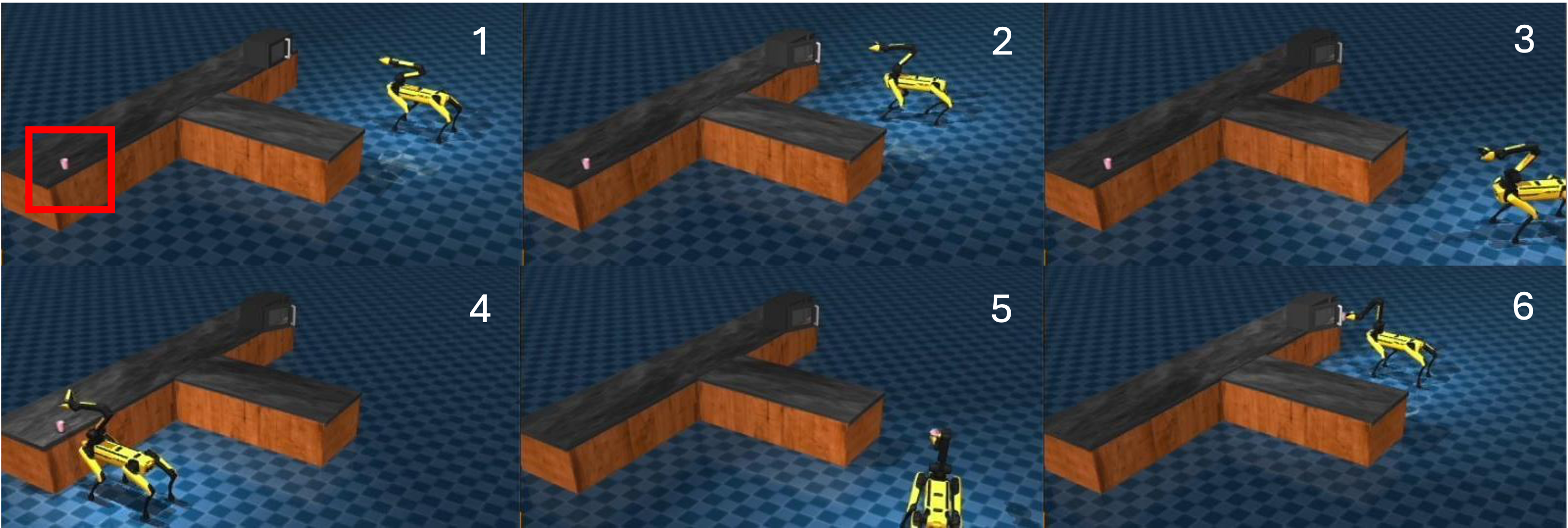}
        \caption{Task 3: Heat-up Tea Long}
        \label{fig:image_c}
    \end{subfigure}
    
    \caption{Illustrations of the simulated mobile manipulation tasks. }
    \label{fig:sim_images}
    \vspace{-5pt} 
\end{figure}

We first evaluate our approach on three simulated mobile manipulation tasks(Fig.~\ref{fig:sim_images}) with the Boston Dynamics Spot robot in MuJoCo~\cite{todorov2012mujoco}: 
\begin{enumerate}
    \item \textbf{Throw Trash:} The robot must walk to a box, pick it up, and throw it into a trash can. The locations of both the box and the trash can are randomized in each episode. Each episode contains around 100 timesteps.
    \item \textbf{Heat-up Tea:} The robot walks to a microwave, opens the door, walks to a cup, picks it up, and uses its body to nudge the microwave door open before placing the cup inside. Each episode contains around 300 timesteps.
    \item \textbf{Heat-up Tea Long:} A more challenging variant of Heat-up Tea where the microwave and the cup are located further apart, increasing the navigation complexity. Each episode contains around 1000 timesteps.
\end{enumerate}
These tasks are designed to be long-horizon and composed of multiple subtasks involving both navigation and manipulation.

The Spot robot is equipped with two cameras: one mounted on the front of the body and another on the end effector. For each task, we collect 400 scripted demonstrations. To enable mobile manipulation behaviors in simulation, we train a Deep Reinforcement Learning (DRL) agent for the low-level locomotion controller. Please see Supplementary Material for more details. 

The low-level policy, denoted as $\pi_{ll}(a_t | s_t, g_t)$, receives proprioceptive information $s_t$ and base velocity commands $g_t$, and outputs joint angle targets for the legs. Arm joint angles are sent directly to the arm’s PD controller. To ensure robust whole-body execution, we randomly sample arm configurations during training so that the robot learns to walk while the arm is in various poses. More details on the low-level controller and the data collection process are provided in the Supplementary Material.

We collect demonstrations such that, at any given time, the robot operates in either locomotion mode or manipulation mode. In locomotion mode, the arm joints are frozen, and only the velocities of the legs are controlled; in manipulation mode, the legs are fixed while the arm is active. In addition to joint and leg velocities, we also record a mode indicator.

During policy training, our network is designed to predict both the joint angle targets for the legs and arm, as well as the mode indicator. During data collection, we apply a sigmoid function to the raw mode signal to generate a smooth, continuous target. This soft target is then used to train the policy, enabling smoother transitions between locomotion and manipulation during execution. During execution, we apply a masking mechanism to activate the leg or arm control outputs based on the predicted mode.

\begin{figure*}[t]
    \centering

    \begin{subfigure}[b]{0.3\textwidth}
        \centering
        \includegraphics[width=\textwidth]{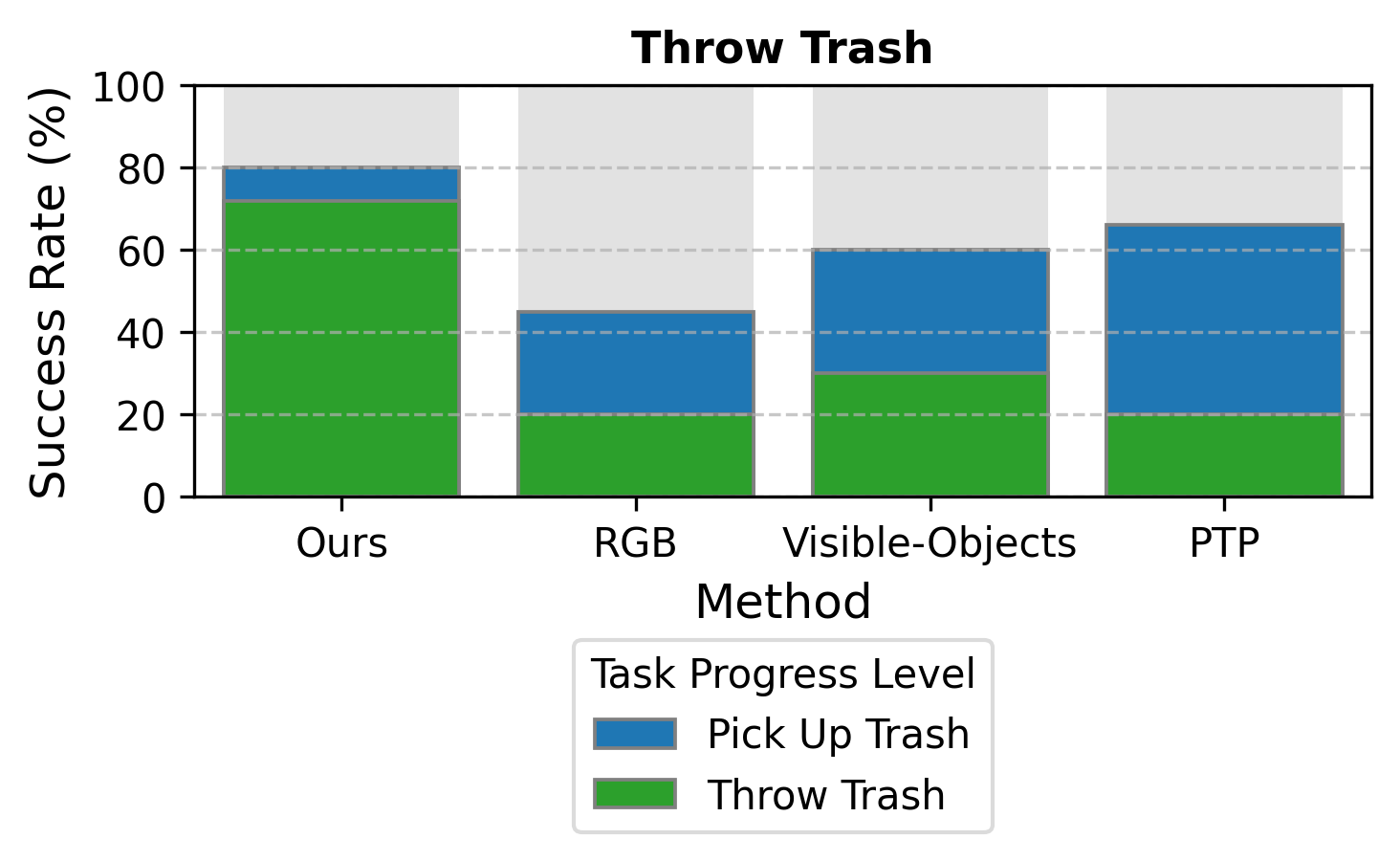}
        \caption{Task 1 Success Rate}
        \label{fig:task1}
    \end{subfigure}
    \hfill
    \begin{subfigure}[b]{0.3\textwidth}
        \centering
        \includegraphics[width=\textwidth]{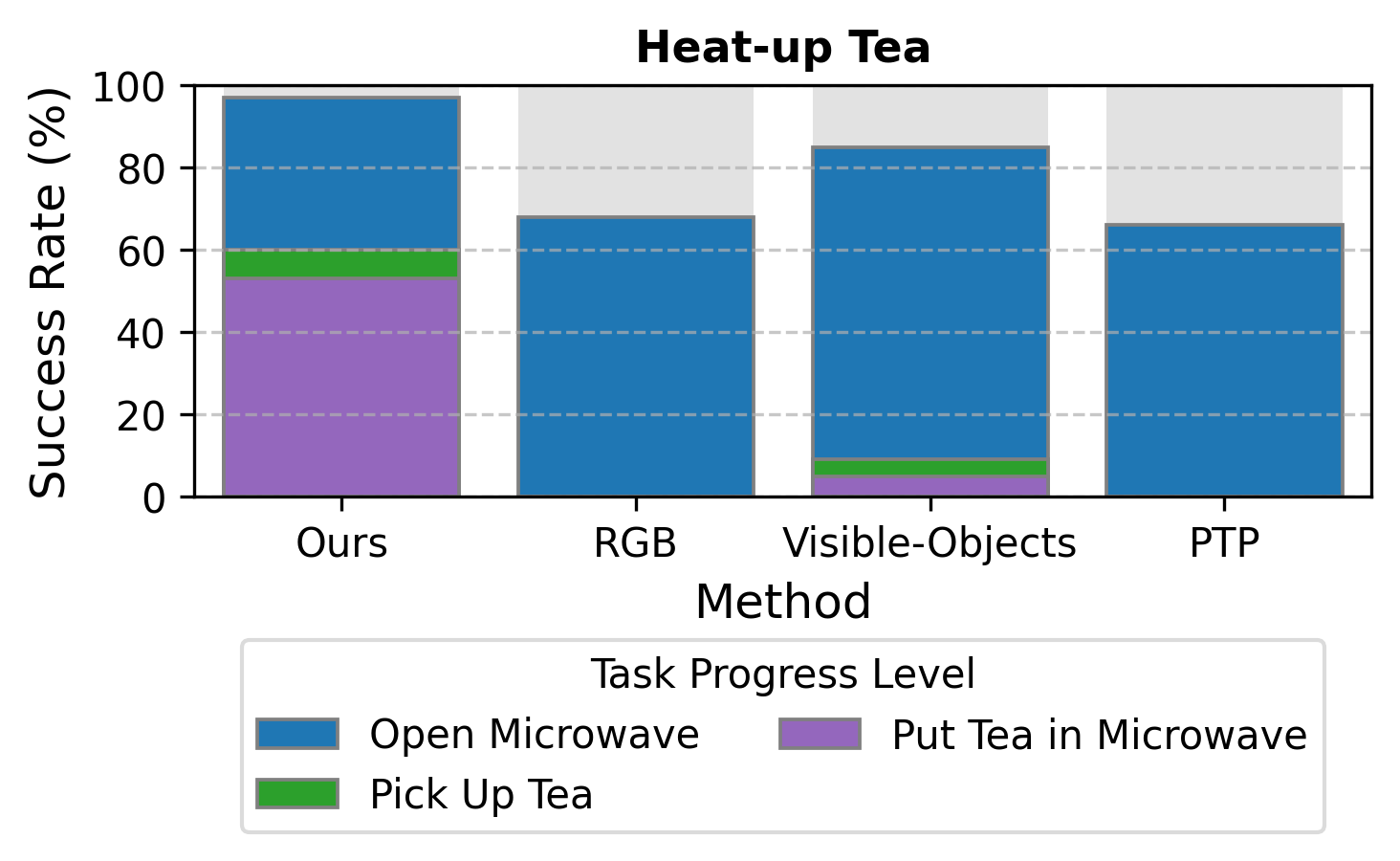}
        \caption{Task 2 Success Rate}
        \label{fig:task2}
    \end{subfigure}
    \hfill
    \begin{subfigure}[b]{0.3\textwidth}
        \centering
        \includegraphics[width=\textwidth]{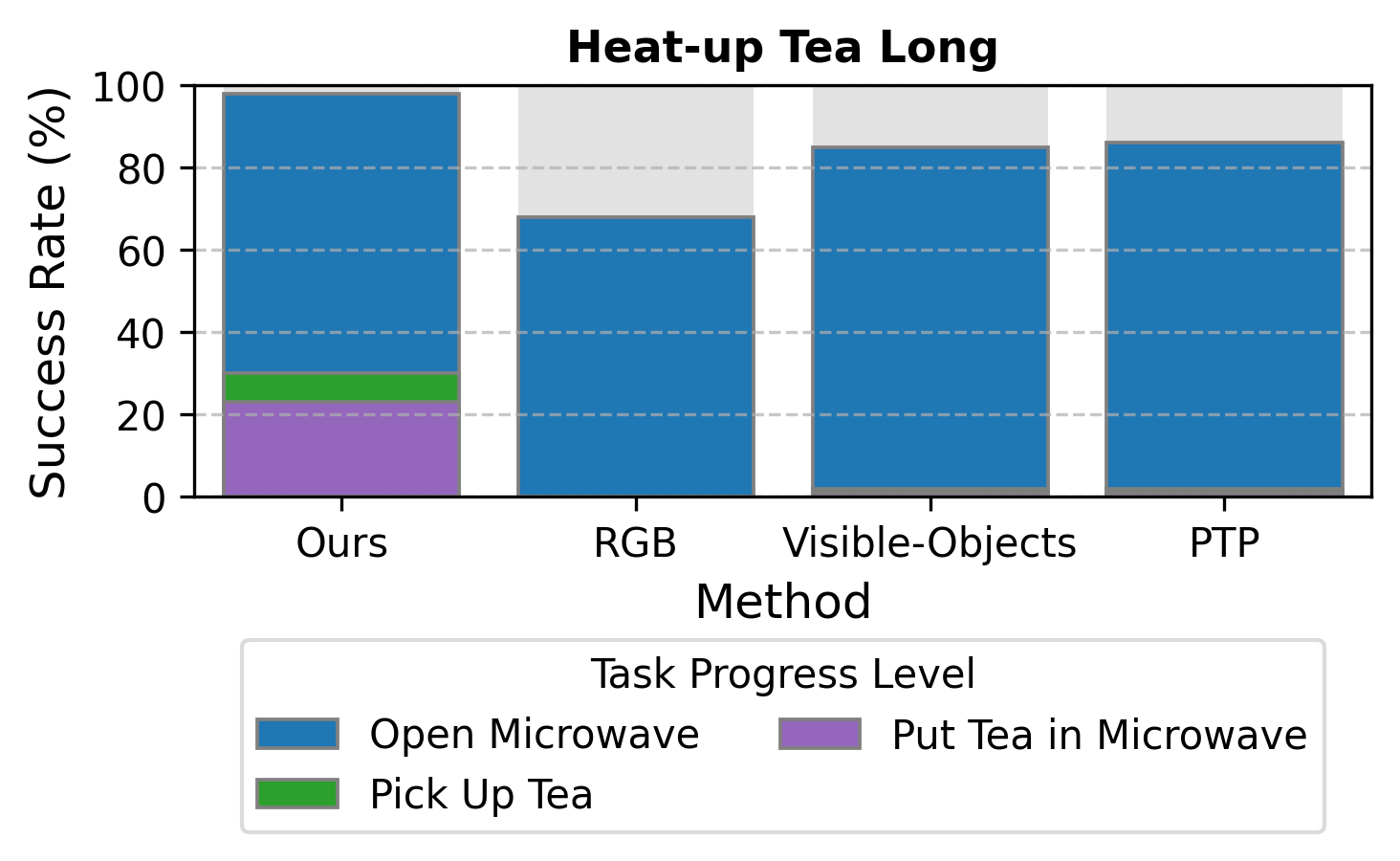}
        \caption{Task 3 Success Rate}
        \label{fig:task3}
    \end{subfigure}

    \vspace{0.5em}

    \begin{subfigure}[b]{0.3\textwidth}
        \centering
        \includegraphics[width=\textwidth]{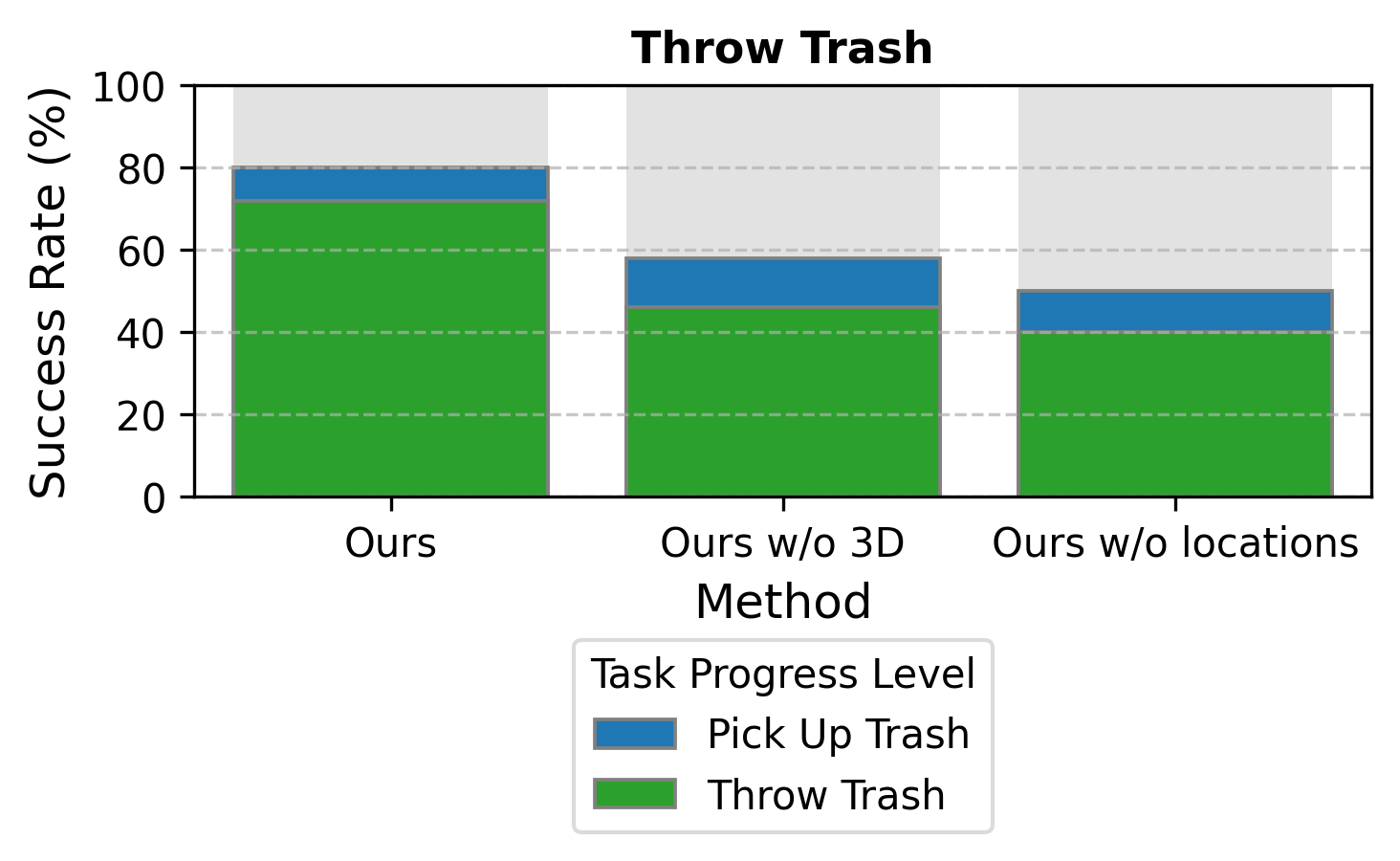}
        \caption{Task 1 Ablation Experiments}
        \label{fig:task4}
    \end{subfigure}
    \hfill
    \begin{subfigure}[b]{0.3\textwidth}
        \centering
        \includegraphics[width=\textwidth]{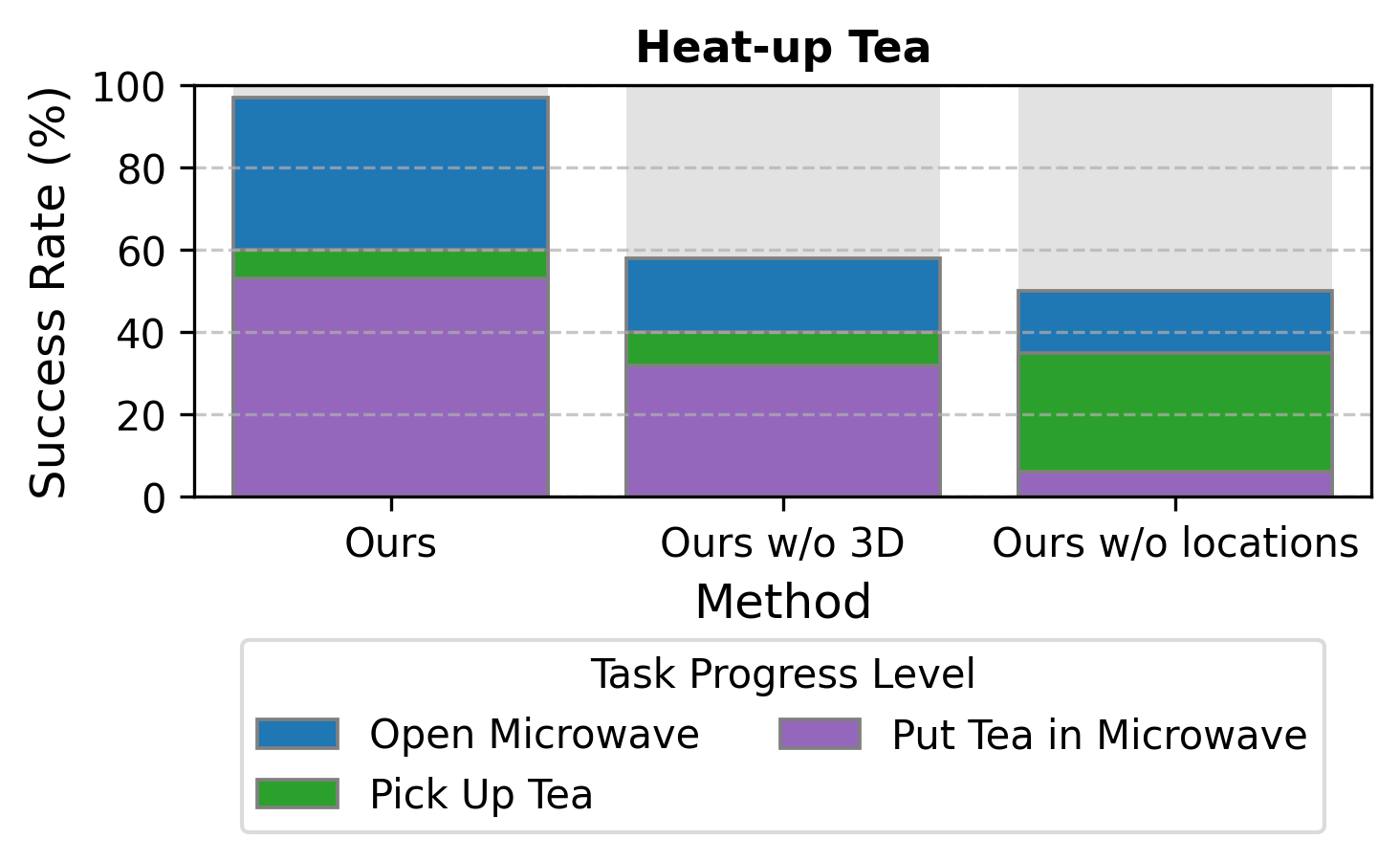}
        \caption{Task 2 Ablation Experiments}
        \label{fig:task5}
    \end{subfigure}
    \hfill
    \begin{subfigure}[b]{0.3\textwidth}
        \centering
        \includegraphics[width=\textwidth]{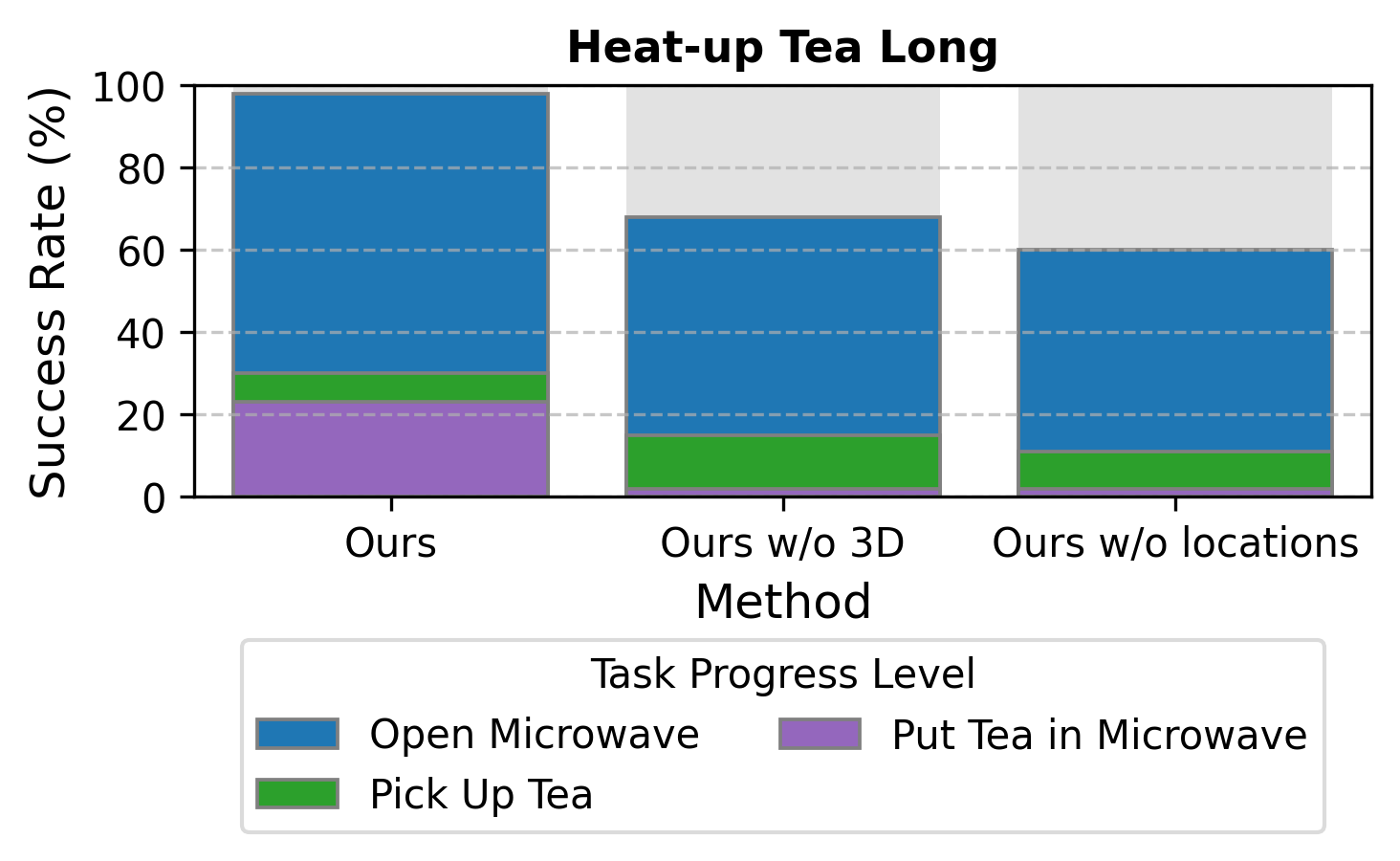}
        \caption{Task 3 Ablation Experiments}
        \label{fig:task6}
    \end{subfigure}

    \caption{
        Success rates of different methods and their ablations across three simulated tasks are represented by stacked bar plots. The lowest bar in each stacked column represents the success rate of that task, while the other bars show success rates for completing earlier subtasks that lead up to the final tasks. The gray bar in the back represents the number of evaluation runs that complete zero subtasks.
    }
    \label{fig:sim_tasks}
    \vspace{-6pt} 
\end{figure*}

\noindent\textbf{Baselines.}
We compare our method with three baselines: 
\begin{enumerate}
    \item \textbf{RGB}: Vanilla diffusion policy that takes in RGB images from two cameras mounted on the Spot and proprioceptive state as input;
    \item \textbf{Visible-Objects}: Diffusion policy that takes instantaneous object-centric representations as input. Compared to our method, these object-centric representations are constructed using only the current camera views.
    \item \textbf{PTP}: To overcome the difficulties of learning long-horizon imitation learning tasks, ~\cite{torne2025learning} proposes a method to input more history frames as well as predict both past and future actions. We implement a variant of diffusion policy where we increase the history frames from 1 to 10 and predict both the past 10 the and future 32 actions.
\end{enumerate} 

\noindent\textbf{Ablations.}
In our scene graph design, each object node $n_i$ encodes not only the visual features of a task-relevant object but also its 2D bounding box coordinates and 3D centroid. To assess the contribution of these components to the overall performance, we conduct a series of ablation studies in addition to the main baselines. Specifically, we evaluate two variants: 
\begin{enumerate}
    \item \textbf{Ours w/o 3D}, which removes the 3D centroid $c_i$ from the node representation. Objects that are currently visible are represented by their object embedding $e_i$ and bounding box $b_i$, and those that are not are represented solely as $e_i$. In other words, there is no memory of \textit{where} in the scene they were observed, just a memory of the fact that they exist in the scene and of their appearance. 
    \item \textbf{Ours w/o object locations}, which removes all location information associated with each object node i.e., each object is represented only by its appearance embedding $e_i$.
\end{enumerate}

\noindent\textbf{Performance Metrics.}
Each simulated task consists of two or three subtasks. To clearly show task progress, we report results using stacked bar plots (Fig.~\ref{fig:sim_tasks}). The bottom segment of each stack indicates the success rate of the full task, while higher segments show success rates for completing earlier subtasks. A gray background bar denotes the fraction of evaluations that fail to complete any subtasks.

\noindent\textbf{Results.}
The results for the simulated tasks are shown in Fig.~\ref{fig:sim_tasks}. We observe that errors in early subtasks often propagate and compound over time, as later stages depend on the resulting scene configuration. This effect is particularly pronounced under partial observability, where recovering from earlier mistakes requires remembering previously observed states. Our method mitigates this issue by maintaining object-centric memory, leading to more consistent performance across long-horizon tasks.

For Heat-up Tea and Heat-up Tea Long in particular, the third step—nudging open the microwave door and placing the cup inside—depends heavily on how the microwave was opened during the very first subtask. Without an explicit memory mechanism, baseline methods struggle to relocate the microwave door hundreds of steps later. The PTP baseline further highlights this limitation: simply increasing the temporal context improves short-horizon reasoning (e.g., on the first subtask) but does not address the fundamental challenges of long-horizon manipulation.

To understand which design choices most contribute to performance, we evaluate two ablations: \textbf{Ours w/o 3D} and \textbf{Ours w/o object locations}. The results in Fig.~\ref{fig:task4}, Fig.~\ref{fig:task5}, and Fig.~\ref{fig:task6} show that removing the 3D centroid or both 3D and 2D location information does not significantly affect performance on the first subtask. However, the success rates drop sharply for the later subtasks and the final task. This suggests that in long-horizon settings, providing explicit spatial information (3D centroids and 2D bounding boxes) is crucial—it helps the policy more reliably localize and track task-relevant objects over time.

\begin{figure}[t]
    \centering
    
    \begin{subfigure}{0.95\linewidth}
        \centering
        \includegraphics[width=\linewidth]{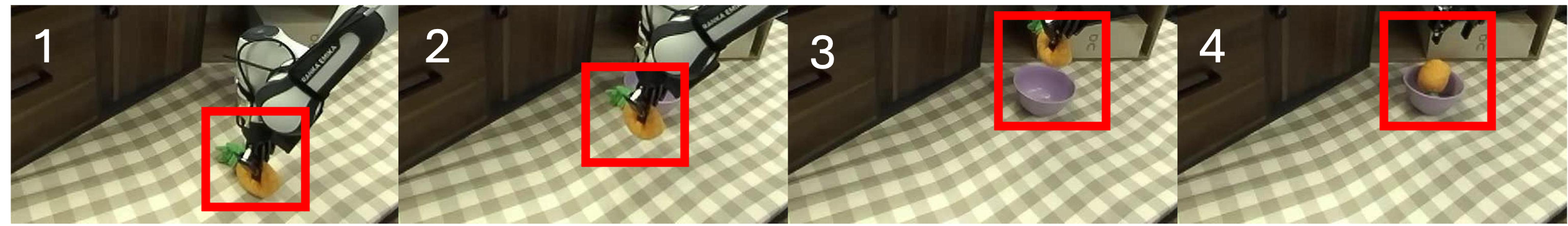}
        \caption{Task 1: Pineapple-Bowl}
    \end{subfigure}
    \vspace{0.5em} 

    \begin{subfigure}{0.95\linewidth}
        \centering
        \includegraphics[width=\linewidth]{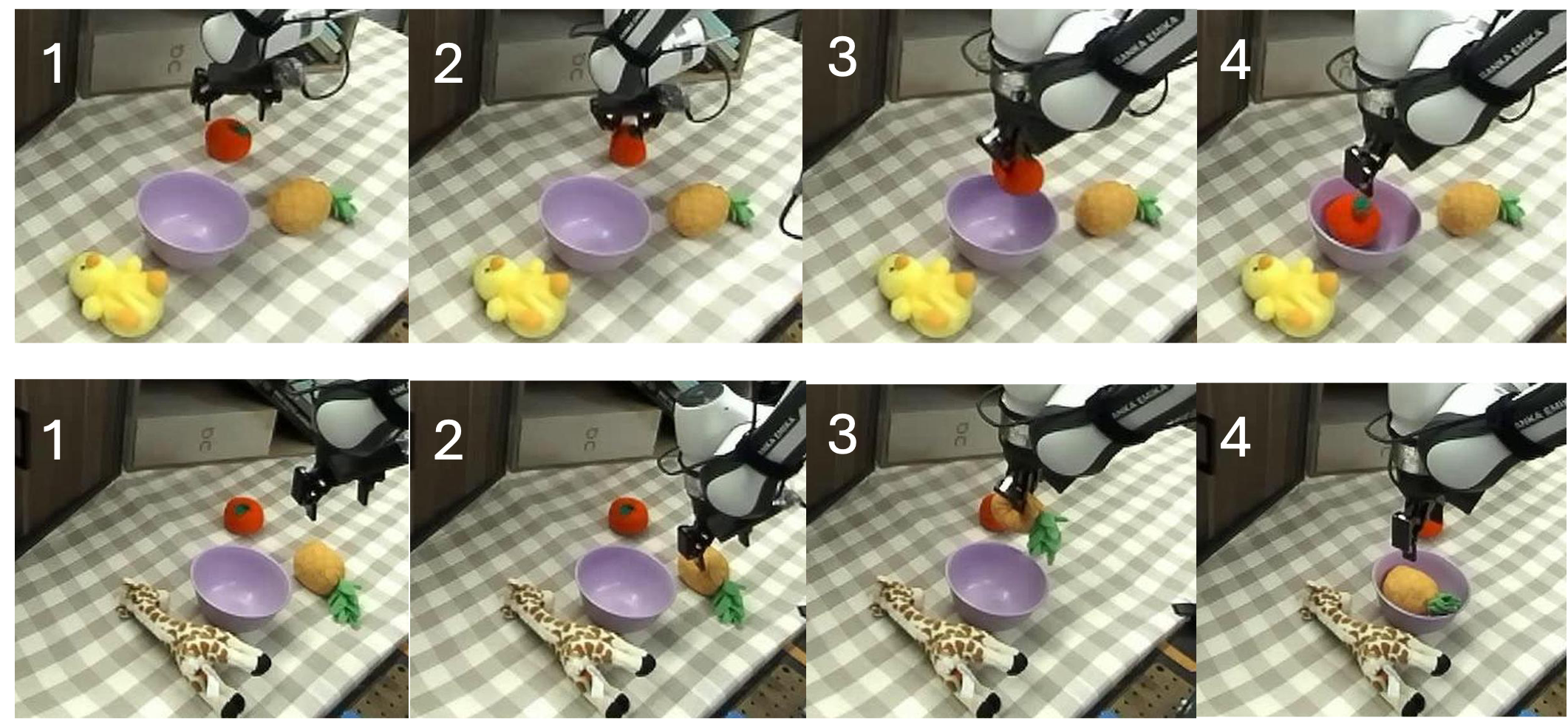}
        \caption{Task 2: Feed-Animals}
    \end{subfigure}
    \vspace{0.5em}

    \begin{subfigure}{0.95\linewidth}
        \centering
        \includegraphics[width=\linewidth]{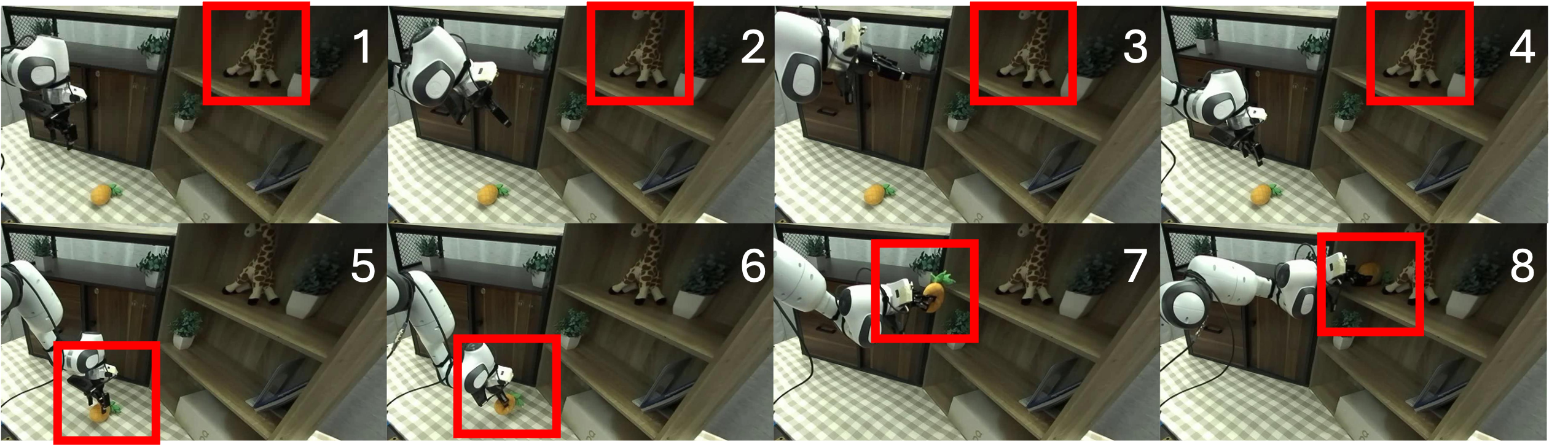}
        \caption{Task 3: Feed-Giraffe}
    \end{subfigure}

    \begin{subfigure}{0.95\linewidth}
        \centering
        \includegraphics[width=\linewidth]{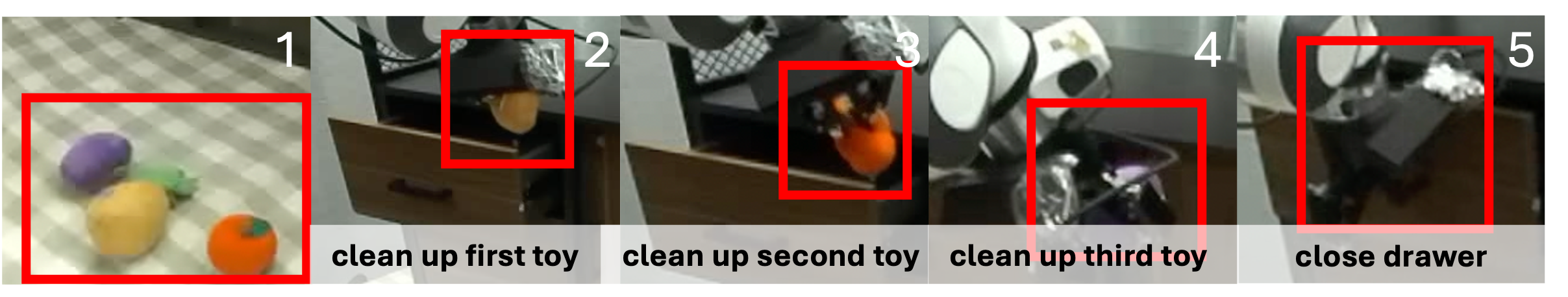}
        \caption{Task 4: Clean-Table}
    \end{subfigure}

    \caption{Illustration of the real-world tabletop manipulation tasks. }
    \label{fig:real_images}
\end{figure}

\begin{figure*}[t]
    \centering
    \includegraphics[width=\textwidth]{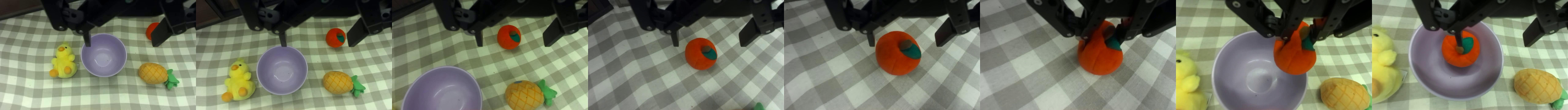}
    
    \caption{\textbf{Partial-Observation Tabletop Setup.} We evaluate our method using a 7-DoF Franka Emika Panda arm with only a wrist-mounted ZED Mini RGB-D camera, removing the side cameras. While this setup simplifies data collection and makes policies invariant to many task-irrelevant scene features, it introduces limited and changing viewpoints. Policies must therefore leverage observations accumulated over the trajectory to reason effectively about the scene.}
    \label{fig:wrist-images}
\end{figure*}

\subsection{Real World Tabletop Manipulation}
Next, we evaluate our method in a partially observed tabletop setup. In these experiments, we use the 7-DoF Franka Emika Panda robotic arm. Unlike the original setup, we remove the two side-mounted cameras, retaining only the wrist-mounted ZED Mini RGB-D camera. 
As we have noted above, this is a commonly used setup in robotic manipulation: several prior works~\cite{Chi2024UniversalMI,Etukuru2024RobotUM} rely solely on wrist-mounted cameras because they permit simpler, morphology-agnostic data collection, and the resulting policies are naturally invariant to many task-irrelevant aspects of the scene because they mostly do not even appear within the image observations. 
However, it also introduces new challenges (see Fig.~\ref{fig:wrist-images}): as the wrist-mounted camera moves during task execution, it often provides limited and changing viewpoints. The policies must therefore rely on observations accumulated throughout the trajectory to reason effectively about the scene.

\begin{figure*}[t]
    \centering

    \begin{subfigure}[b]{0.3\textwidth}
        \centering
        \includegraphics[width=\textwidth]{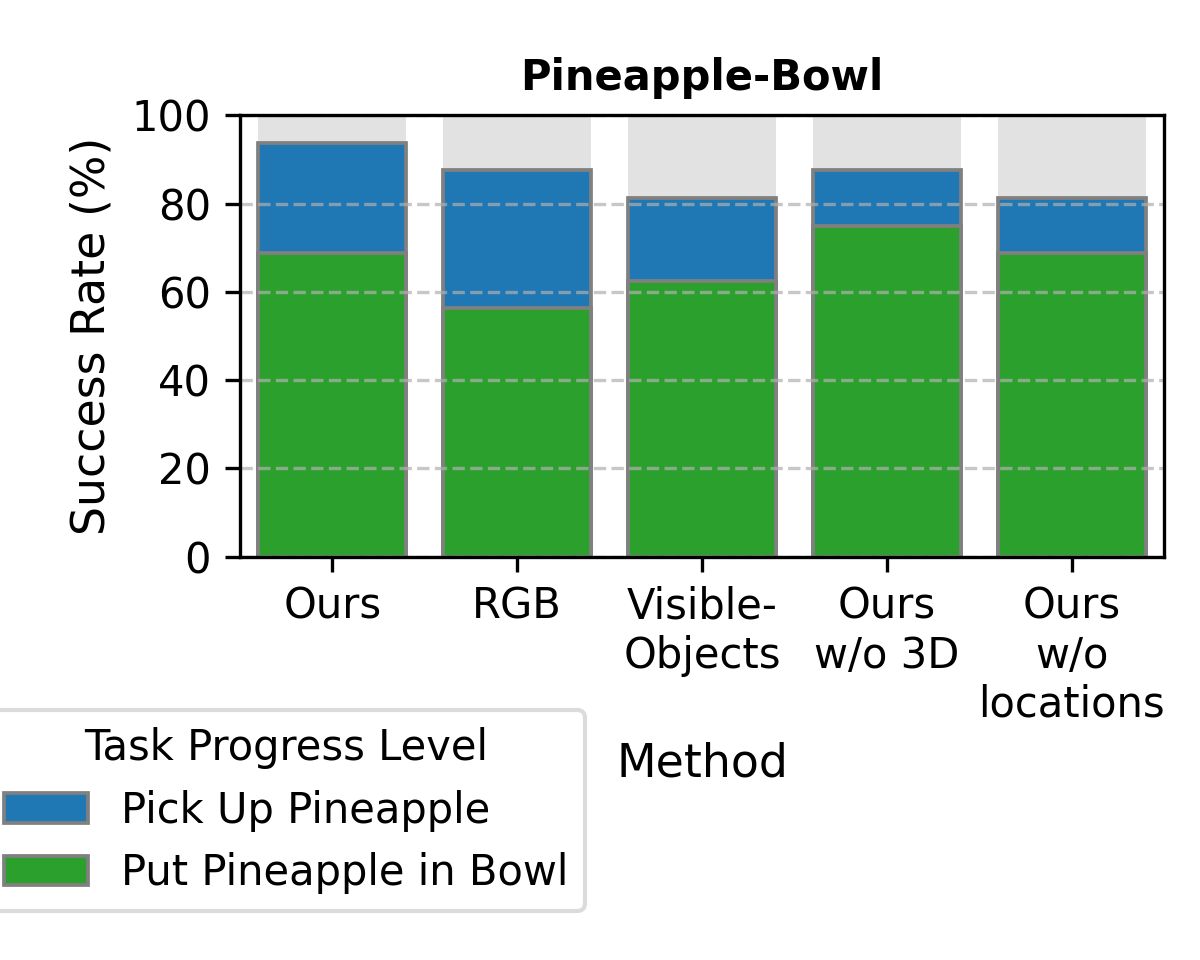}
        \caption{Task 1 Success Rate}
        \label{fig:task1}
    \end{subfigure}
    \hfill
    \begin{subfigure}[b]{0.3\textwidth}
        \centering
        \includegraphics[width=\textwidth]{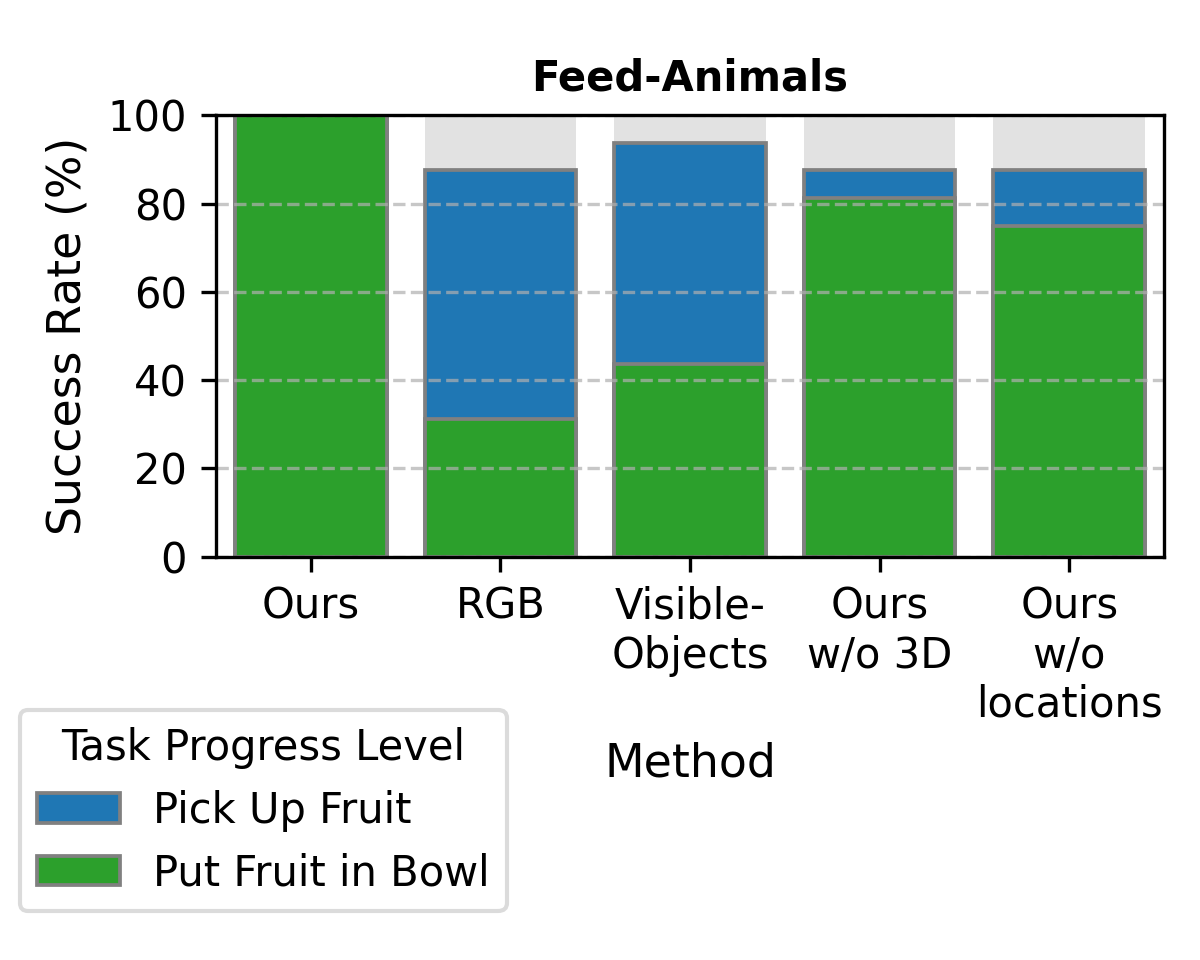}
        \caption{Task 2 Success Rate}
        \label{fig:task2}
    \end{subfigure}
    \hfill
    \begin{subfigure}[b]{0.3\textwidth}
        \centering
        \includegraphics[width=\textwidth]{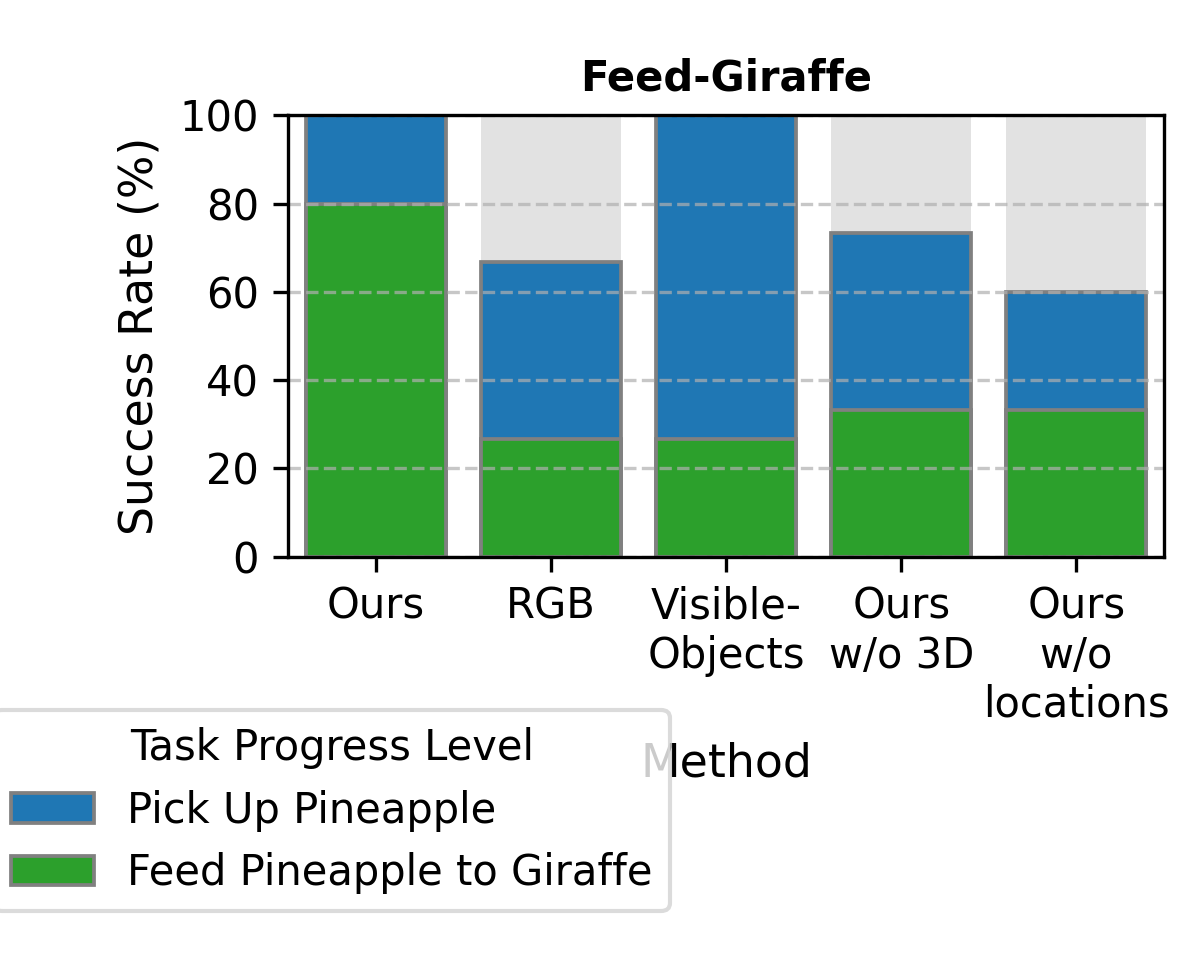}
        \caption{Task 3 Success Rate}
        \label{fig:task3}
    \end{subfigure}
    \hfill
    \begin{subfigure}[b]{0.3\textwidth}
        \centering
        \includegraphics[width=\textwidth]{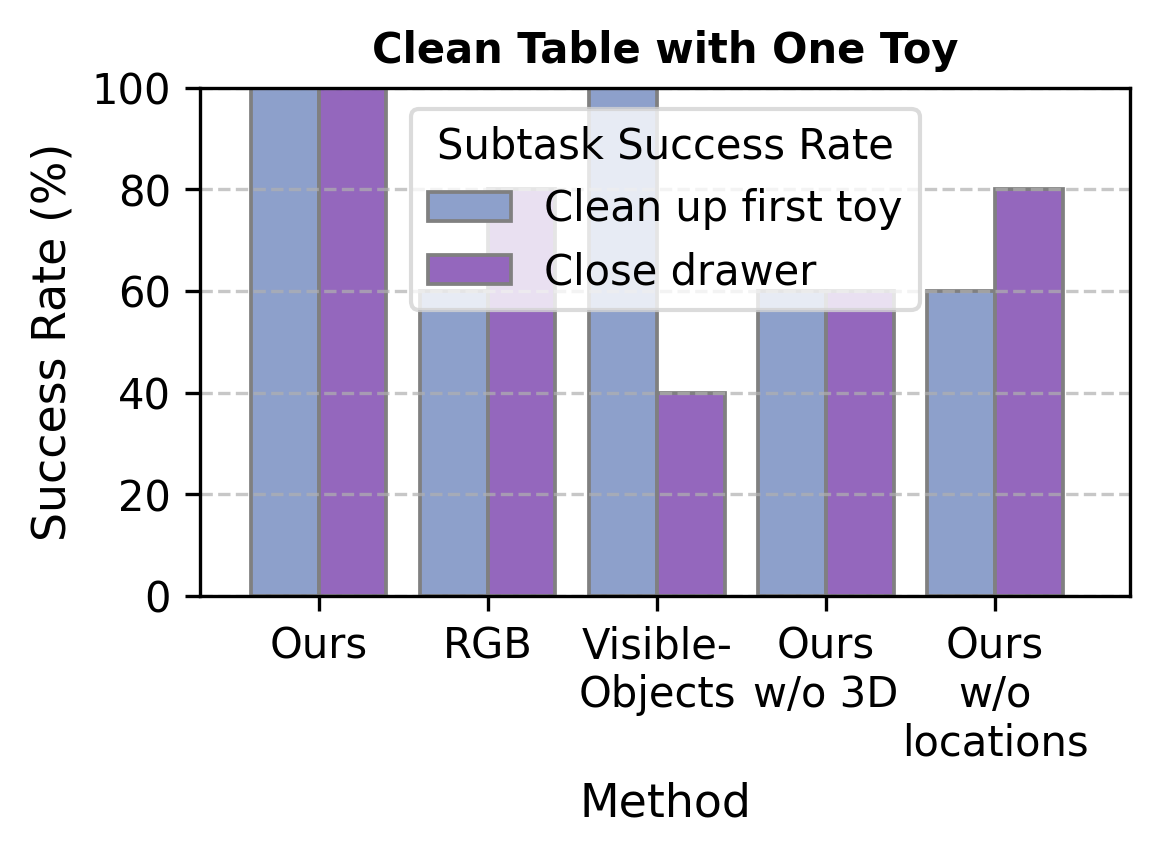}
        \caption{Task 4 Scenario 1 Success Rate}
        \label{fig:task3}
    \end{subfigure}
    \hfill
    \begin{subfigure}[b]{0.3\textwidth}
        \centering
        \includegraphics[width=\textwidth]{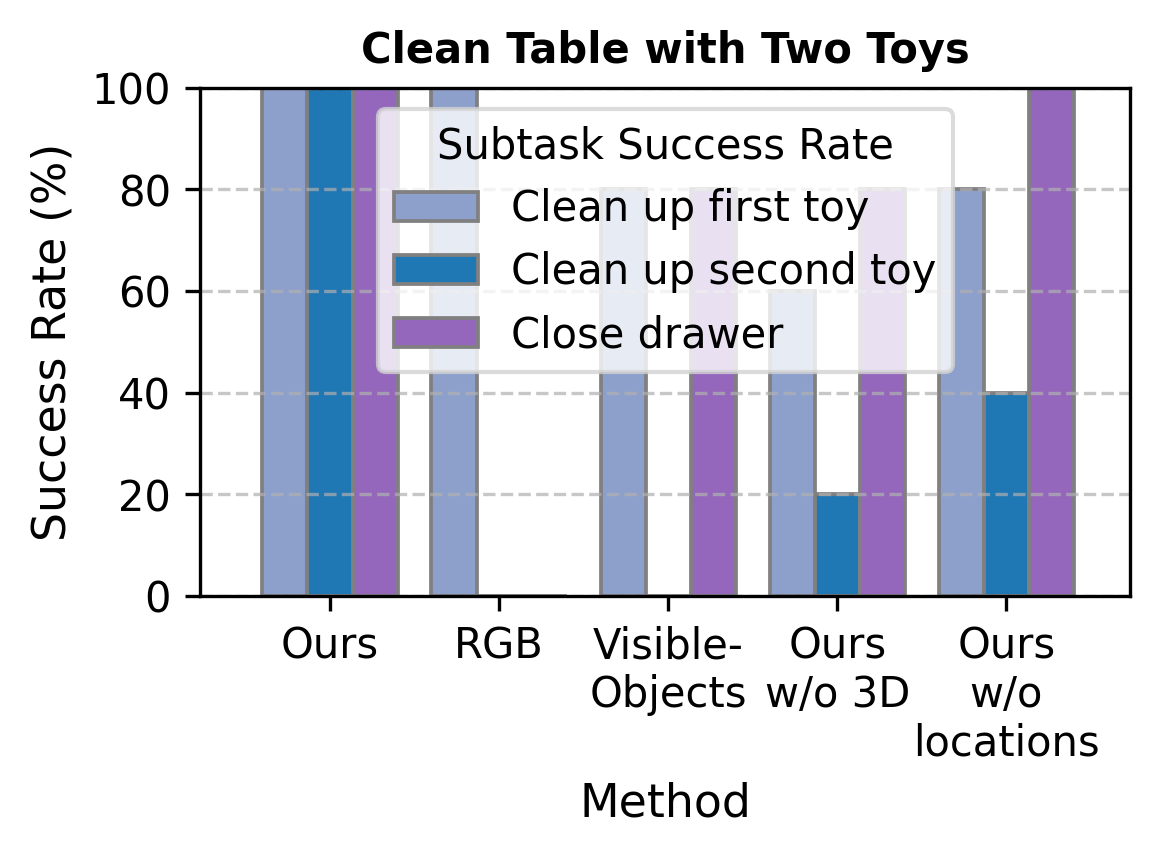}
        \caption{Task 4 Scenario 2 Success Rate}
        \label{fig:task3}
    \end{subfigure}
    \hfill
    \begin{subfigure}[b]{0.3\textwidth}
        \centering
        \includegraphics[width=\textwidth]{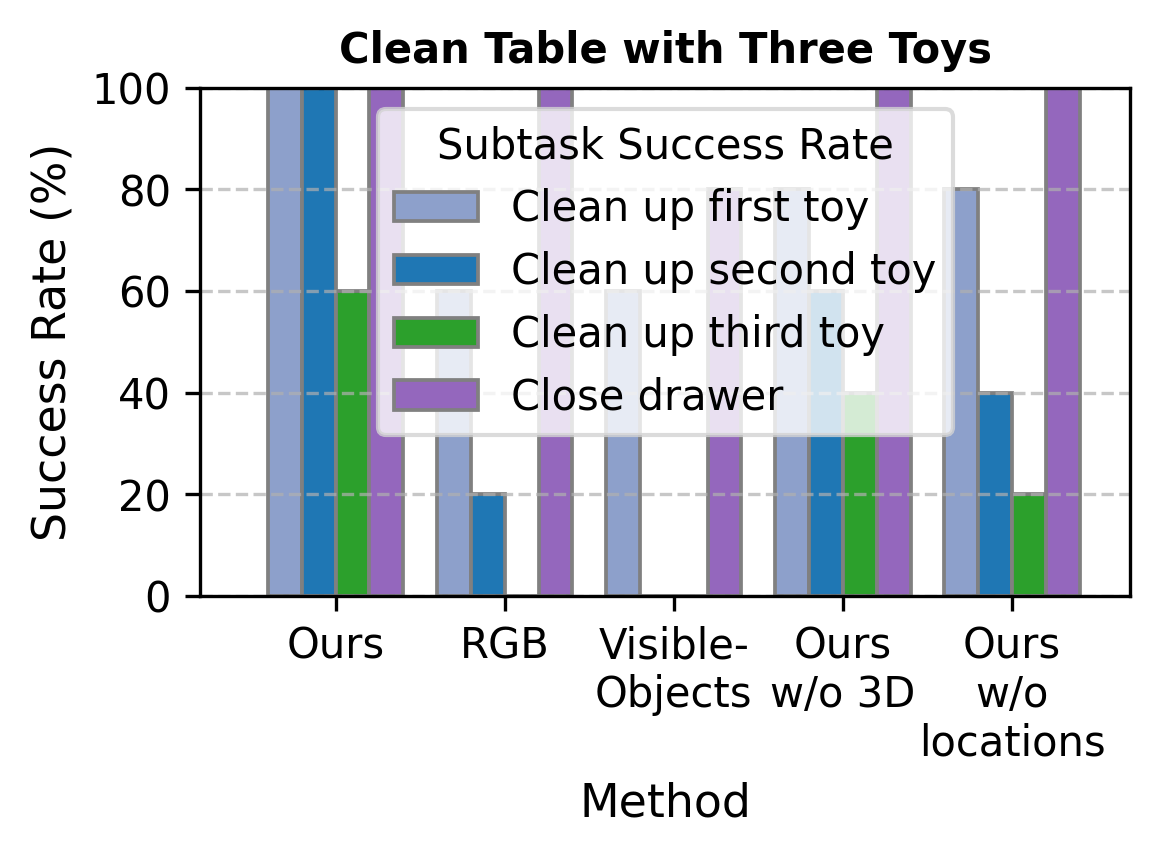}
        \caption{Task 4 Scenario 3 Success Rate}
        \label{fig:task3}
    \end{subfigure}
    \caption{
        Success rates of different methods and their ablations on the first three real-world tabletop manipulation tasks are shown as stacked bar plots. Each column corresponds to a method, and each segment represents the success rate of a subtask. The bottom segment denotes the success rate of completing the full task, while higher segments indicate success on intermediate subtasks. The gray segment represents runs that fail to complete any subtask. For the fourth task, we report results separately for three scenarios: cleaning up one, two, or three toys.
    }
    \label{fig:real_tasks} 
    \vspace{-10pt} 
\end{figure*} 
\noindent\textbf{Data Collection and Setup.}
We set up a tabletop environment that contains a variety of everyday objects and several cabinets mounted on the countertop. A ZED Mini RGB-D camera is mounted on the robot’s wrist and serves as the sole visual sensor, providing RGB-D image streams for both data collection and policy learning. For each task, as illustrated in Fig.~\ref{fig:real_images}, we collect 300 teleoperated demonstrations. The task suite is designed to increase in difficulty progressively, allowing us to evaluate the method’s ability to handle partial observability and long-horizon challenges. Additional implementation and environment details are provided in the Supplementary Material. Our tasks are:
\begin{enumerate}
\item Put Pineapple in Bowl \textbf{(Pineapple-Bowl):} The robot must pick up a plush pineapple and place it into a bowl. The main challenge is partial observability: with a wrist-mounted camera, the bowl leaves the field of view once the pineapple is lifted, requiring the policy to maintain a memory of its location.

\item Feed Two Animals \textbf{(Feed-Animals):} A pineapple and an apple are placed on the table, and either a giraffe or a chicken appears at the start of the episode. The robot must choose the correct fruit based on the animal (pineapple for giraffe, apple for chicken), testing its ability to retain and act on initial visual context.

\item Feed Giraffe \textbf{(Feed-Giraffe):} A giraffe is positioned on one of three bookshelf levels. The robot must first scan the shelf to locate it, then pick up a pineapple and place it at the corresponding level. This requires long-horizon reasoning and persistent memory of the target location.

\item Clean Table \textbf{(Clean-Table):} One to three fruit toys are placed on the table. The robot must collect all objects and place them into a drawer, and only close the drawer once the table has been fully cleared. This requires tracking which objects have already been moved and ensuring the policy does not prematurely terminate the task.

\end{enumerate}

\noindent\textbf{Results.} See Fig.~\ref{fig:real_tasks} for quantitative results. We compare our method against all baselines and ablations, excluding PTP because it rarely progresses beyond the first task stage in simulated tasks.

For the \textbf{Pineapple-Bowl} task, our method achieves the highest success rate in picking up the pineapple and performs on par with \textbf{Ours w/o 3D} and \textbf{Ours w/o object locations} for placing it in the bowl, as it maintains a consistent memory of the bowl’s location even when being out of view. \textbf{Visible-Objects} outperforms \textbf{RGB}, indicating that object-centric representations improve grasping and placement accuracy, consistent with prior OCR-based work~\cite{Shi2024ComposingPO,Zhu2023LearningGM}. However, both \textbf{RGB} and \textbf{Visible-Objects} frequently misplace the pineapple when the bowl is far from its initial position.

For the \textbf{Feed-Animals} task, we observe a large performance gap between our method and the baselines. Both \textbf{RGB} and \textbf{Visible-Objects} tend to pick up a random fruit regardless of the target animal, reflecting an inability to retain and act on initial visual context under partial observability.

\textbf{Feed-Giraffe} is a challenging task, requiring the robot to scan the bookshelf, locate the giraffe, and place the pineapple at the correct level. This long-horizon dependency makes it difficult for memory-less policies: while both our method and \textbf{Visible-Objects} can pick up the pineapple, baselines often either fail to retrieve it or lose track of the giraffe’s location during placement.

For the \textbf{Clean-Table} task, the robot must remove all objects before closing the drawer, requiring it to keep track of which items have already been cleared. Policies without explicit memory often prematurely close the drawer or leave objects behind, indicating difficulty in maintaining task progress over time. In contrast, our method achieves more reliable completion by maintaining a persistent representation of object states throughout the episode.

\textbf{Error Propagation and Runtime Analysis.}
The above results suggest that performance is primarily limited by long-horizon reasoning rather than perception. To better understand system behavior, we analyze error sources and runtime.

\textbf{Failure Analysis.}
Our system combines LLM-based task parsing, object segmentation, tracking, and policy inference. The LLM is queried once per task to identify relevant objects, which are segmented in the first frame using Grounded-SAM and tracked with XMem. In simulation, we observe no failures from perception or tracking. In real-world experiments (64 episodes), segmentation fails in 4 cases due to occlusion or ambiguous boundaries. We observe no failures from LLM-based object enumeration or from tracking after initialization. Instead, most failures arise from the imitation policy in long-horizon settings, consistent with the challenges observed in tasks such as \textbf{Feed-Giraffe} and \textbf{Clean-Table}.

\textbf{Runtime.}
Expensive components are used sparingly: the LLM is queried once per task, and GroundingDINO runs once per episode (0.53 s). During execution, mask propagation takes 0.19 s per frame. The diffusion policy runs at 3.5-3.8 Hz on a real robot using a single NVIDIA 3090 GPU (around 260–285 ms per step), enabling real-time closed-loop control. The primary runtime cost during execution is mask propagation, while other components are amortized over the episode.

\section{Limitations and Conclusions}
We introduce task-relevant scene graphs as a compact, dynamically updated memory for imitation learning under partial observability. By encoding only relevant objects, the representation enables reasoning over past observations beyond the robot’s field of view. While effective for long-horizon tasks, it depends on accurate object identification and may struggle in highly cluttered or dynamic environments. This suggests scene graphs as a promising interpretable memory for future vision-language-action models.

%% file: sec/5_suppl.tex
\clearpage
\setcounter{page}{1}
\maketitlesupplementary

\section{Additional Information about Our Task-Driven Scene Graph}
\label{sec:supp-graph}
\begin{figure}[htbp]
    \centering 
\begin{subfigure}{0.4\textwidth}
  \includegraphics[width=\linewidth]{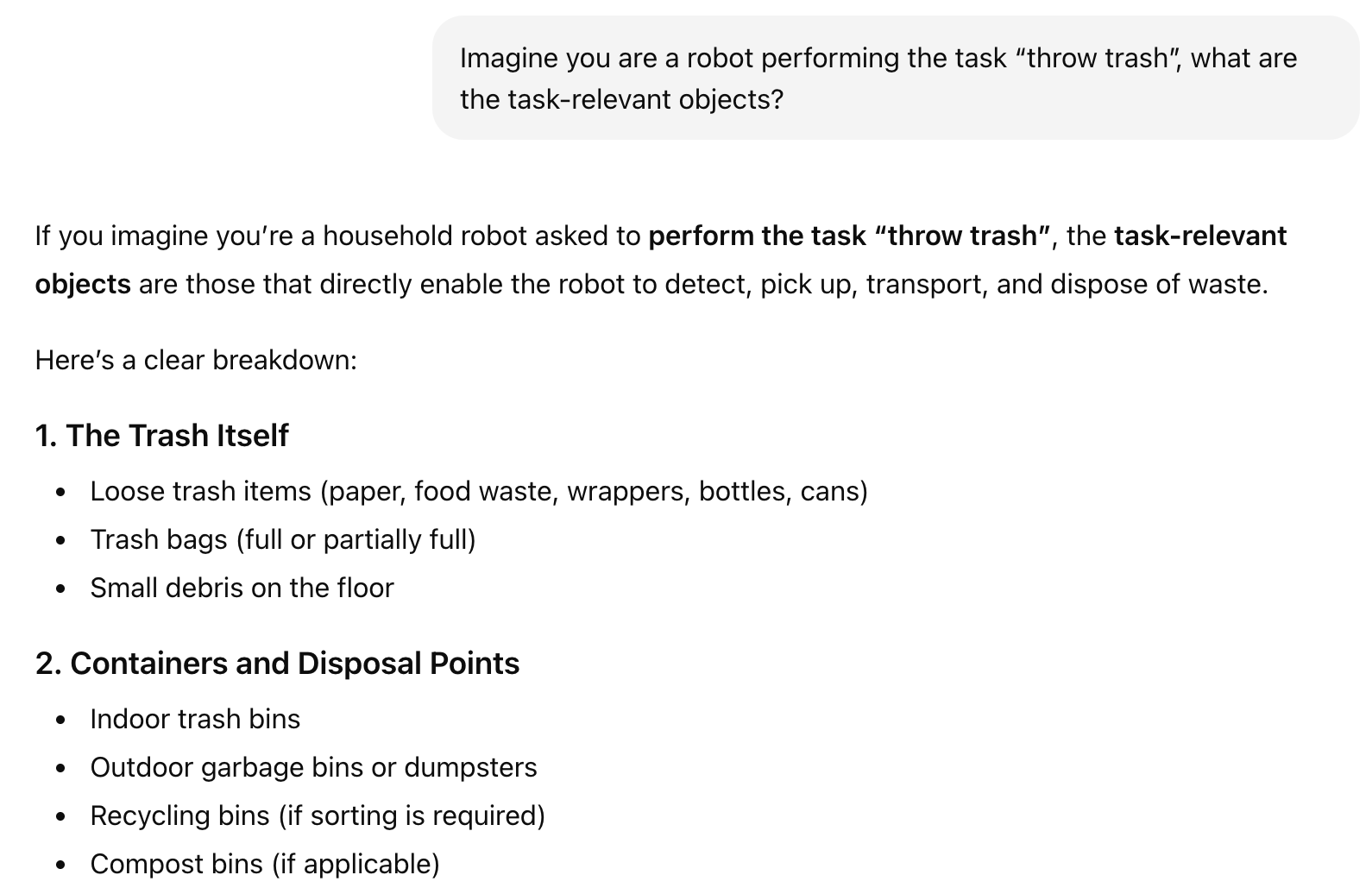}
  \caption{Throw trash}
  \label{fig:1}
\end{subfigure}\hfil 
\begin{subfigure}{0.4\textwidth}
  \includegraphics[width=\linewidth]{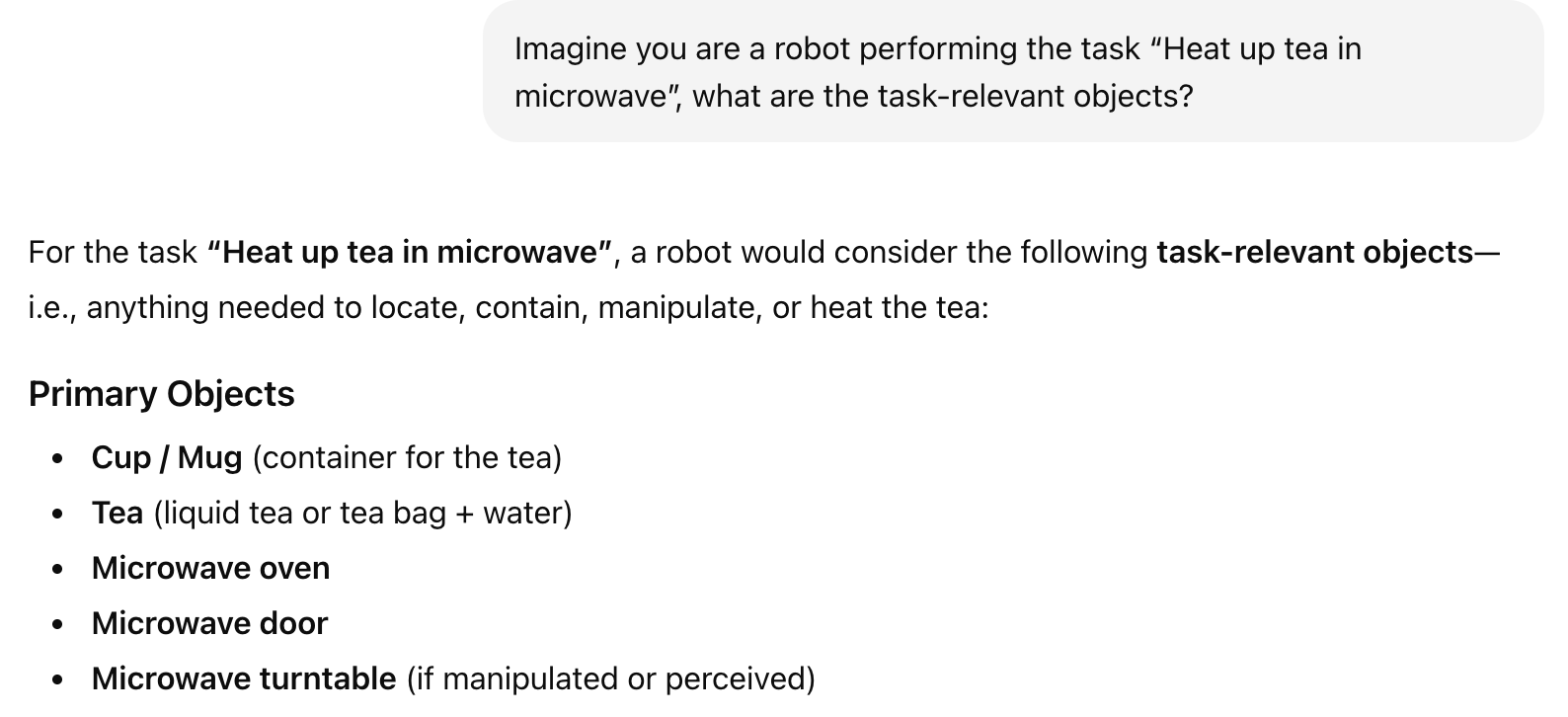}
  \caption{Heat up tea in the microwave}
  \label{fig:2}
\end{subfigure} 
\medskip
\begin{subfigure}{0.4\textwidth}
  \includegraphics[width=\linewidth]{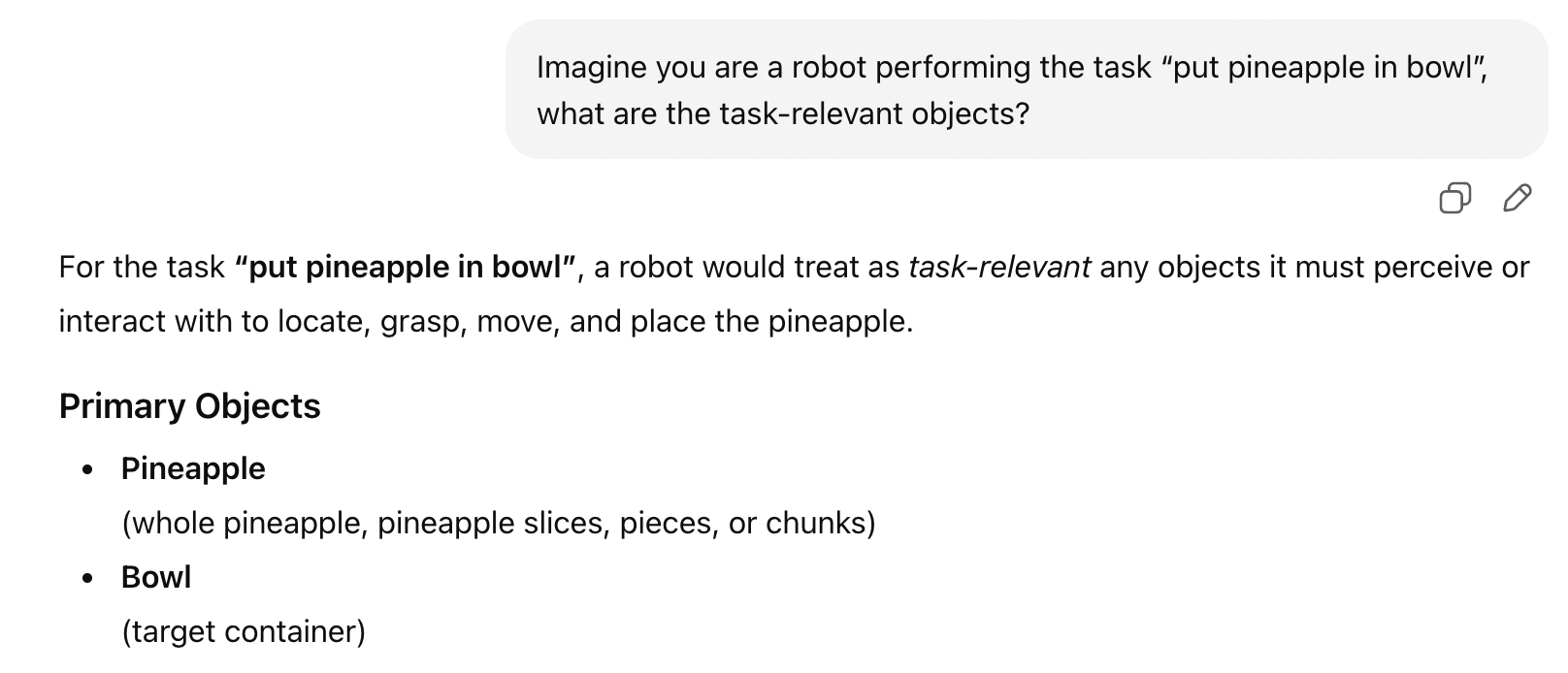}
  \caption{Put pineapple in bowl}
  \label{fig:3}
\end{subfigure}\hfil
\begin{subfigure}{0.4\textwidth}
  \includegraphics[width=\linewidth]{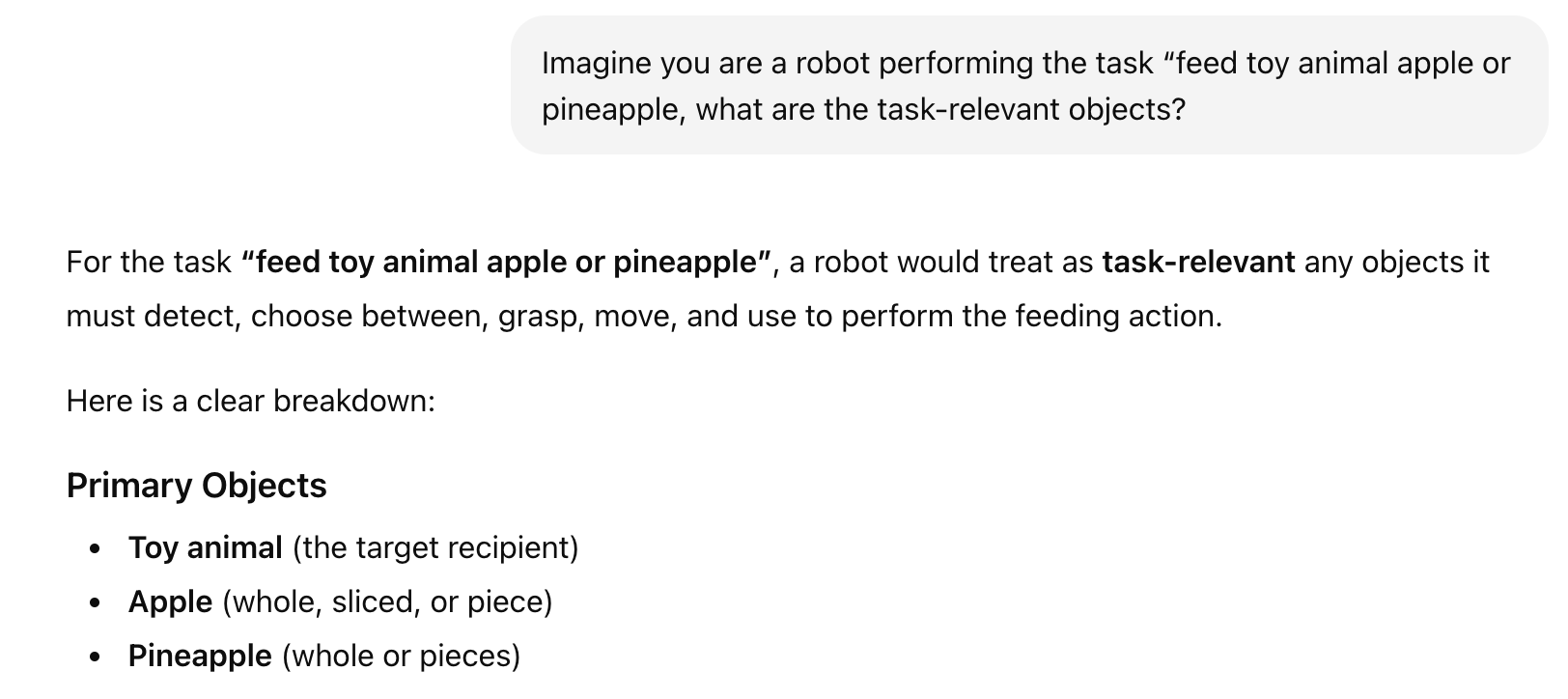}
  \caption{Feed toy animal apple or pineapple}
  \label{fig:4}
\end{subfigure} 
\medskip
\begin{subfigure}{0.4\textwidth}
  \includegraphics[width=\linewidth]{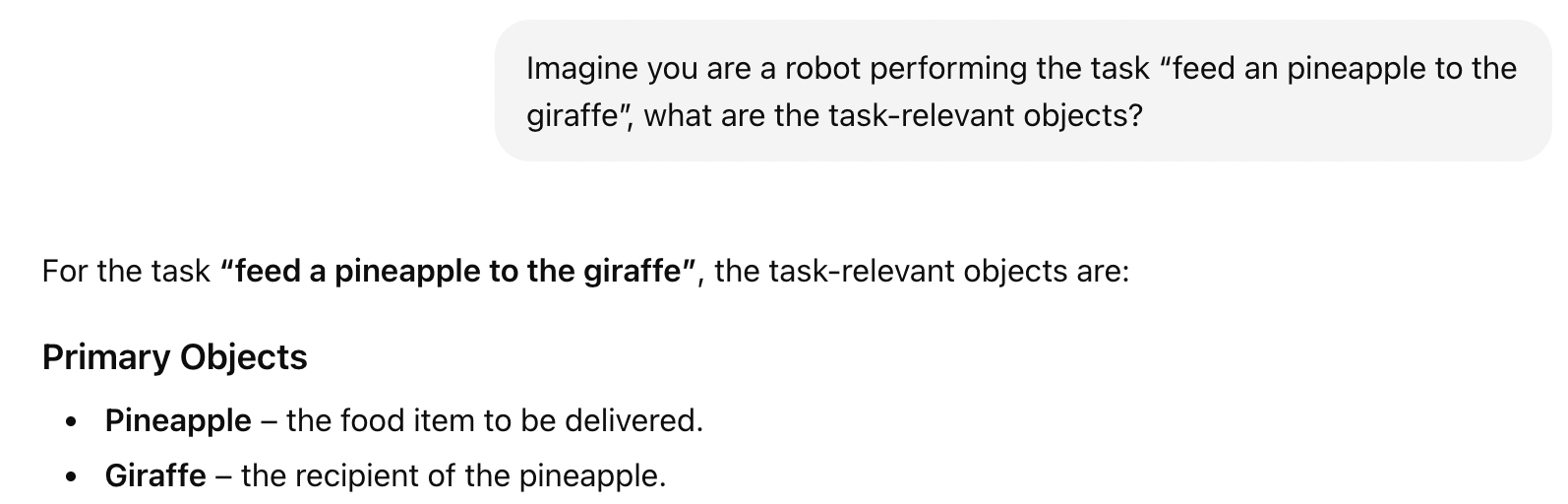}
  \caption{Feed pineapple to giraffe}
  \label{fig:5}
\end{subfigure}\hfil 
\medskip
\begin{subfigure}{0.4\textwidth}
  \includegraphics[width=\linewidth]{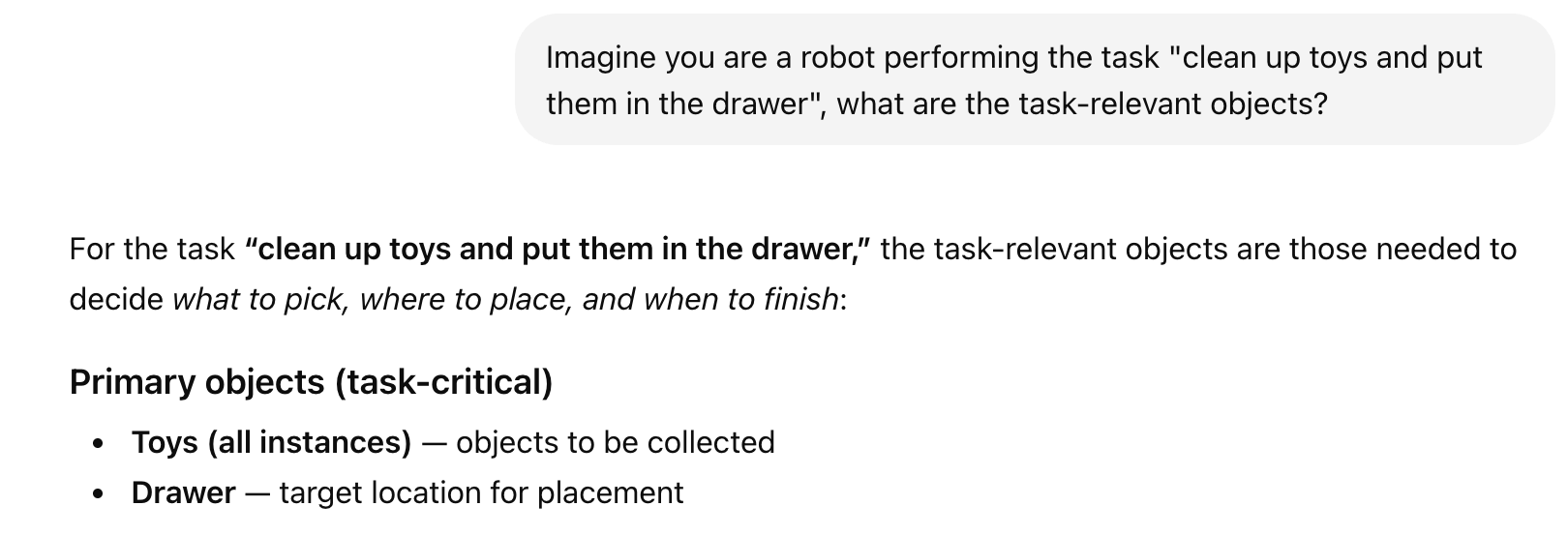}
  \caption{Clean Table}
  \label{fig:6}
\end{subfigure}\hfil 
\caption{GPT-4 responses for task-relevant object name identification}
\label{fig:gpt4}
\end{figure}
Concretely, our scene graph is implemented as a \emph{two-level tree} in which the root node is represented by the CLS token extracted from the DINO-v2 encoder applied to the current image observation. The second level of the tree consists of the set of task-relevant object nodes described in Section 3.1. Although this structure lacks the deeper hierarchical layers (e.g. room $\rightarrow$ objects), we find that a shallow, two-tier topology is fully sufficient for the mobile manipulation tasks considered in this work, where the primary need is to reason about objects with respect to the robot’s current view. Importantly, this formulation remains fully compatible with richer multi-level hierarchies as explored in CLIO~\cite{Maggio2024ClioRT} and HODOR~\cite{hodor}; future extensions of our system could incorporate room-level or task-semantic nodes by simply adding additional parent layers above our current root node.

From the perspective of the downstream transformer encoder, the parent (root) node is always supplied as the \emph{first token} in the sequence. Consistent with standard transformer architectures, we attach a positional embedding to each node token, including both the root and all object nodes. This positional encoding gives the model an explicit mechanism to distinguish the structural role of each node and allows the transformer to reliably identify the root as the global context provider. As a result, even in the absence of explicit relational edges, the model learns to interpret the object nodes as children conditioned on a shared scene-level parent, enabling effective reasoning over this minimal yet structured two-level scene graph.
\paragraph{Prompt and Entity Lists for Our Tasks}
In this paper, we use natural task descriptions to prompt a large language model(LLM) for a list of relevant objects. Identifying the right entities (e.g., “microwave” and “cup”) given task prompts (e.g., “Heat up a cup of tea”) is very easy for modern LLMs, so we used very simple prompts: “Imagine you are a robot performing the task “$\langle$insert-task-name$ \rangle$”, what are the task-relevant objects?” With GPT-4, this works with 100\% accuracy across all our tasks. We include the results from GPT-4 for all tasks in Fig~\ref{fig:gpt4}(a)-(f).

\section{Technical Details for Collecting Demonstrations in Simulated Mujoco Environment}
\paragraph{Low-Level Controller.}
During demonstration collection, we train a locomotion controller in simulation. Following the MDP formulation of~\cite{rudin2022learning}, we employ Proximal Policy Optimization (PPO)~\cite{schulman2017proximal} within IsaacLab~\cite{mittal2025isaac} to learn a robust quadruped walking policy.

The resulting low-level controller, denoted $\pi_{ll}(a_t \mid s_t, g_t)$, receives proprioceptive observations $s_t$ along with a desired base velocity command $g_t$, and outputs joint-angle targets for the legs. The arm joints are controlled separately via a PD controller. To improve the robustness of the learned locomotion behavior, we randomize the arm configuration during training, enabling the robot to walk stably even when the arm is placed in arbitrary poses.
\paragraph{Mode Switching.}
During data collection, the robot operates in one of two regimes: \emph{locomotion mode} and \emph{manipulation mode}. In locomotion mode, the arm joints are frozen and $\pi_{ll}$ controls only the leg velocities; in manipulation mode, the base remains stationary while the arm executes joint-space actions. Along with joint states and leg velocities, we record a mode indicator that specifies the robot's operational regime.

To avoid abrupt discontinuities at switching boundaries---e.g., when the robot transitions between navigating and manipulating---we encode the mode using a temporally smoothed signal. For each mode transition occurring at time $t$, we generate a transition-specific contribution
\[
\tilde{m}^{(t)}_k =
\begin{cases}
\ \ \sigma(k - t), & \text{if switching into manipulation}, \\
-\sigma(k - t), & \text{if switching into locomotion},
\end{cases}
\]
where $\sigma$ denotes the logistic sigmoid. The final mode value at timestep $k$ is obtained by summing all contributions from past transitions:
\[
m_k = \sum_{t \in \mathcal{T}} \tilde{m}^{(t)}_k,
\]
where $\mathcal{T}$ denotes the set of all transition times. This produces a smooth, continuous, and bidirectional encoding of the robot's operational regime, stabilizing training and avoiding artifacts caused by instantaneous mode flips.

During training, the high-level policy predicts this smoothed mode value $m_k$ jointly with all other arm and leg action outputs, enabling the policy to learn coordinated transitions between locomotion and manipulation behaviors. During test time, we convert the predicted mode value into a discrete operational mode via thresholding: if $m_k > 0.5$, the robot is considered to be in manipulation mode; otherwise, it is treated as being in locomotion mode.

\section{Technical Details for Real World Tabletop Manipulation Experiments}
We utilize a 7-DoF Franka robotic arm operating under a continuous joint-control action space at 15 Hz. A ZED Mini camera is mounted on the robot’s wrist, and the captured wrist images are encoded using DINO-v2. The resulting CLS token is incorporated as an additional input to the policy during training. Operating under velocity control, the robot’s action space comprises a 6-DoF joint velocity vector alongside a single gripper command representing open or close actions, yielding a 7-dimensional continuous action output from the policy. All policies are trained for 200 epochs using a single NVIDIA 3090 GPU. For each task, we collected 300 demonstrations. Qualitative comparisons across the three tasks can be found in the supplementary materials, which include a presentation containing demonstration videos for each method.

\newpage